\documentclass[a4paper, 10 pt, conference]{ieeeconf}

\IEEEoverridecommandlockouts
\usepackage{cite}
\usepackage{amsmath,amssymb,amsfonts}
\usepackage{algorithmic}
\usepackage{graphicx}
\usepackage{textcomp}
\usepackage{xcolor}
\usepackage{gensymb}

\usepackage{caption}
\usepackage{subcaption}

\usepackage{algorithm}

\usepackage{array}

\def\BibTeX{{\rm B\kern-.05em{\sc i\kern-.025em b}\kern-.08em
    T\kern-.1667em\lower.7ex\hbox{E}\kern-.125emX}}

\begin{document}

\title{General Anomaly Detection of Underwater Gliders \\
Validated by Large-scale Deployment Datasets}

\author{Ruochu Yang, Chad Lembke, Fumin Zhang, and Catherine Edwards
\thanks{
The research work is supported by ONR grants  N00014-19-1-2556 and N00014-19-1-2266;  AFOSR grant FA9550-19-1-0283; NSF grants GCR-1934836,  CNS-2016582 and ITE-2137798; and NOAA grant NA16NOS0120028.
}
\thanks{Ruochu Yang and Fumin Zhang are with the School of Electrical and Computer Engineering, Georgia Institute of Technology, Atlanta, USA. Chad Lembke is with the College of Marine Science, University of South Florida, St. Petersburg, USA. Catherine Edwards is with the Skidaway Institute of Oceanography, University of Georgia, Savannah, USA}
}

\maketitle

\begin{abstract}
    Underwater gliders have been widely used in oceanography for a range of applications. However, unpredictable events like shark strikes or remora attachments can lead to abnormal glider behavior or even loss of the instrument. This paper employs an anomaly detection algorithm to assess operational conditions of underwater gliders in the real-world ocean environment. Prompt alerts are provided to glider pilots upon detecting any anomaly, so that they can take control of the glider to prevent further harm. The detection algorithm is applied to multiple datasets collected in real glider deployments led by the University of Georgia's Skidaway Institute of Oceanography (SkIO) and the University of South Florida (USF). In order to demonstrate the algorithm generality, the experimental evaluation is applied to four glider deployment datasets, each highlighting various anomalies happening in different scenes. Specifically, we utilize high resolution datasets only available post-recovery to perform detailed analysis of the anomaly and compare it with pilot logs. Additionally, we simulate the online detection based on the real-time subsets of data transmitted from the glider at the surfacing events. While the real-time data may not contain as much rich information as the post-recovery one, the online detection is of great importance as it allows glider pilots to monitor potential abnormal conditions in real time.
\end{abstract}

\section{Introduction}
\label{intro}

Underwater gliders are extensively used in ocean research for activities such as ocean sampling, surveillance, and other purposes \cite{doi:10.1080/00207170701222947, 4476150, hou2020, nicholson2008present, schofield2007slocum}. However, given the complexity of the ocean environment and the long-duration of glider missions, unexpected events such as shark attacks, wing loss, or attachment of marine species can cause gliders to operate abnormally or even totally fail \cite{stanway2015white, mccosker2006white, baehr2013swimming}. In such cases, the gliders may drift to unexpected areas, making localization and rescue operations challenging. Furthermore, it can be difficult to detect the abnormal behavior of gliders, particularly when external disturbances arise, due to the lack of monitoring devices \cite{gertler2017fault, chen2012robust, aslansefat2014strategy, wang2020active}. The deployment of monitoring devices for gliders or the addition of self-monitoring of performance would increase mission costs and pilot complexity. Typically, glider pilots can only rely on heavily subsetted data transmitted by the glider in real time to form hypotheses about potential anomalies. Sometimes, they just resort to climb and dive ballast data to assess if the glider is surfacing or diving as expected. However, this empirical detection can never be conclusively confirmed as the mission is going on. To address this challenge, we develop an anomaly detection algorithm that systematically utilizes simple glider data such as glider speed, heading, and trajectory. This algorithm is feasible for theoretical validation on numerous real-world glider datasets, and runs autonomously in real-time, as opposed to manual detection by human pilots. By monitoring gliders in real-time, the algorithm allows glider pilots to take appropriate actions promptly to ensure the safety and success of missions.

Different strategies have been in the field of underwater robotics to identify abnormal behavior of underwater gliders. Some anomaly detection algorithms focus on changes in robot motion, such as roll angle or pitch angle, to detect possible motion deviation or a foreign object attached to the glider \cite{7404462, anderlini2020autonomous}. Some algorithms monitor the power consumption or motor performance of the glider, as variations in these parameters can indicate degeneration of individual components, such as propellers and rotors  \cite{fagogenis2016online, sun2016thruster, caiti2015enhancing}. Other algorithms utilize machine learning techniques to identify anomalous behavior by analyzing sensor data collected by the glider over time, such as changes in the speed, roll, pitch, or depth \cite{asalapuram2019novel, wu2021anomaly, bedja2022smart}. However, most of the existing research relies on shore-based manual implementations and does not resolve issues like inability to perform online detection on the gliders or lack of real-time experimental verification. In addition, it is essential to determine whether the detected anomaly is false positive \cite{chandola2009anomaly, yassin2013anomaly}. When the ocean current speed is significantly greater than the maximum speed of the marine robot, it can lead to a considerable performance degradation. Under such circumstances, false alarms should be avoided since the anomaly caused by an unexpected ocean current is unrelated to the glider itself. In practice, it is challenging to separate flow speed and glider speed due to hardware limitations, but leveraging the Controlled Lagrangian Particle Tracking (CLPT) framework \cite{szwaykowska2017controlled}, the anomaly detection algorithm in \cite{cho2021learning} generates real-time estimates of the glider speed and flow speed from the trajectory and heading angles. The estimated glider speed is compared with the normal speed range to detect anomalies, while the algorithm-estimated flow speed is compared with the glider-estimated flow speed to avoid false alarms.

We initially validate the anomaly detection algorithm by using two real-life deployment datasets \cite{yang2023anomaly}. Building upon this previous work, we aim to extend the algorithm to large-scale datasets, thus effectively handling various anomalies in diverse missions. We also plan to simulate online implementation of the detection to enable real-time interaction with glider pilots. This objectives constitute primary motivation of this paper, and our main contribution are summarized as follows.
\begin{itemize}
    \item We demonstrate generality of the anomaly detection algorithm based on four glider datasets collected in real deployments featuring diverse anomalies. 

    \item We simulate online mode implementation of the algorithm to a real glider deployment with limited data streams in real time for the first time.
\end{itemize}

The SouthEast Coastal Ocean Observing Regional Association (SECOORA) glider Franklin, operated by Skidaway Institute of Oceanography (SkIO), and the University of South Florida (USF) gliders USF-Sam, USF-Gansett, and USF-Stella provide numerous examples of valuable experimental data in which anomalies may be associated with marine bio-hazards. Promising anomaly detection results of these datasets are shown to well match glider pilots' hindcast analysis. Building off its efficacy, the real-time anomaly detection algorithm is incorporated into the autonomous glider navigation software GENIoS\_Python \cite{yang2023real} to better assist human pilots as an add-on warning functionality.

This paper is organized as follows. Section \ref{anomaly detection algo} illustrates the framework of the anomaly detection algorithm. Section \ref{experimental evalution} describes the experimental setup of glider deployments, verifies the algorithm by detecting anomalies in large-scale real experiments, and simulates the online implementation on subsetted glider datasets. Section \ref{conclusion} provides conclusions and future work.

\section{Anomaly Detection Algorithm}
\label{anomaly detection algo}

\begin{figure}[ht]
        \centerline{\includegraphics[width=0.48\textwidth]{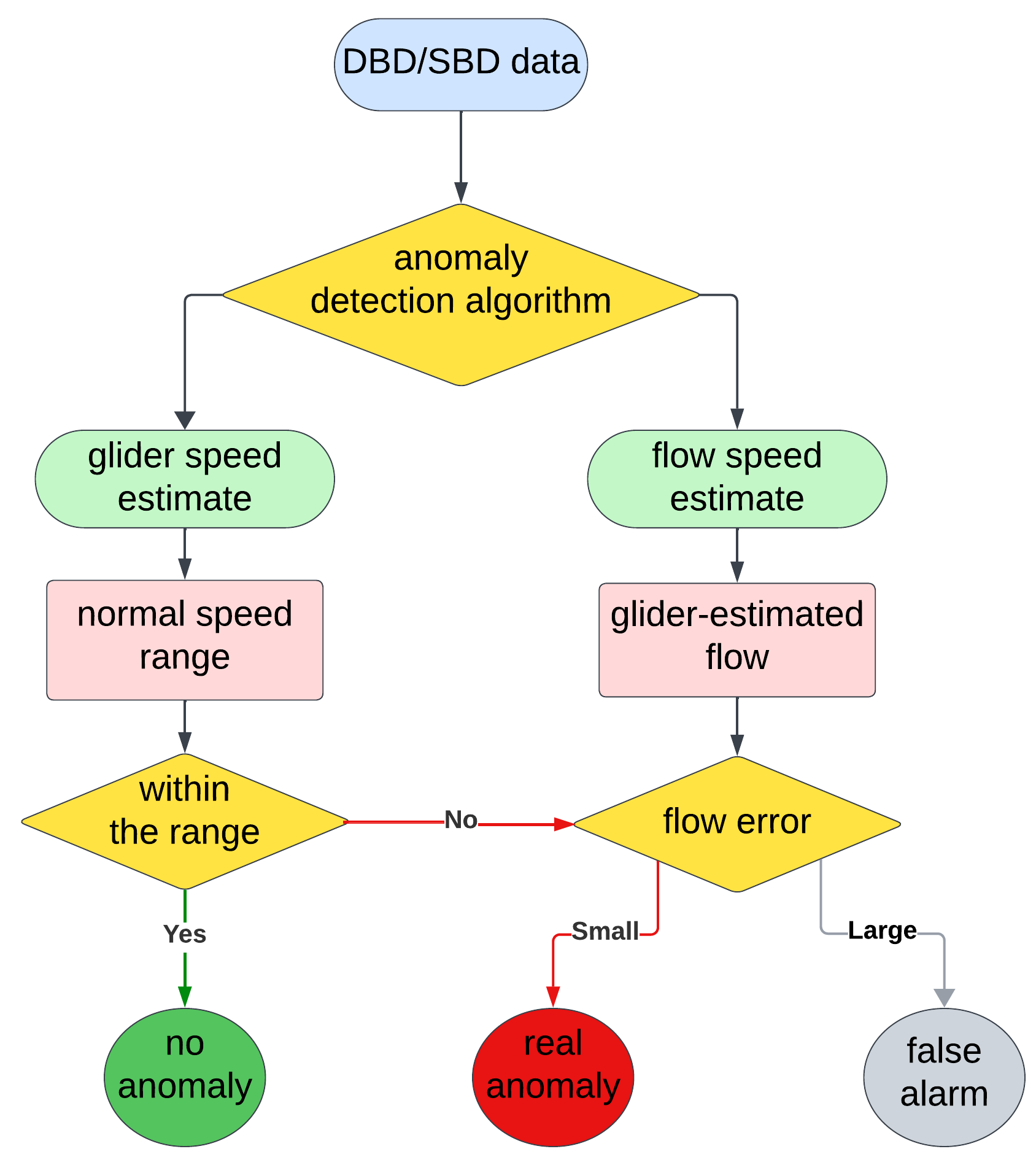}}
        \caption{Pipeline of anomaly detection and false alarm elimination. Three steps: 1) generate glider speed estimate and flow speed estimate based on DBD or SBD data; 2) compare glider speed to detect anomalies; 3) compare flow speed for false alarm.}
        \label{pipeline}
    \end{figure}

The pipeline of anomaly detection and false alarm elimination is shown in Fig.~\ref{pipeline}. By generating the glider speed estimate, the algorithm assumes no anomaly if the estimate is within the normal speed range. Otherwise, the glider may be encountering issues. The flow speed estimate is checked against the glider-estimated flow speed to circumvent any false alarm. 
    
We model the glider dynamics as follows:
\begin{gather}
    \dot{x} = F_R(x, t)  + V_R(t) \Psi_c(t) \nonumber \\
    \Psi_c(t)  = [\cos \psi_c(t), \sin\psi_c(t)]^T,
\label{eq 1}
\end{gather}
where $F_R$ is the true flow field, $x$ is the true glider position, $V_R$ is the true glider speed, and $\psi_c$ is the true heading angle. As shown in \eqref{eq 4}, we model the ocean flow field by spatial-temporal basis functions \cite{liang2012real},  
\begin{gather}
    F_R(x, t) = \theta \phi(x,t)
    \label{eq 4}
\end{gather}
where $\theta$ is the unknown parameter to be estimated,
\begin{gather}
    \phi = \begin{bmatrix} \phi^1(x,t) & \cdots & \phi^N(x,t) \end{bmatrix}^T\\
    \phi^i(x, t) = \exp^{-\frac{||x - c_i||}{ 2 \sigma_i}} \cos(\omega_i t + \upsilon_i)
\end{gather} are the basis functions, $c_i$ is the center, $\sigma_i$ is the width, $\omega_i$ is the tidal frequency, $\upsilon_i$ is the tidal phase, and $N$ is the number of basis functions. Using the heading $\Psi_c(t)$ and the true trajectory $x(t)$, the detection algorithm can generate estimates of the glider speed $V_R$ and  the unknown  flow parameter $\theta$. These estimates will converge to true values as long as the maximum trajectory estimation error (CLLE) converges to zero. The heading data $\Psi_c(t)$ and the glider trajectory $x(t)$ are always available from full post-recovery Dinkum binary data (DBD) and typically included in subsetted short Dinkum binary data (SBD) sent in real-time. Three gains $K$, $\Bar{\gamma}$ and $s$ are designed to accelerate the estimating process. If the estimated glider speed falls within an expected range $V_L(t) \in [V_{min}, V_{max}]$, where the maximum glider speed $V_{max}$ and the minimum glider speed $V_{min}$ are defined a priori, no anomaly should have happened. Otherwise, the glider may not be operating normally. Additionally, introduce $F_L(t)$ as the algorithm-estimated flow speed, and $F_M(t)$ as the glider-estimated flow speed which can be generated by ocean models or sensor measurements. By defining a criteria $p_E$, we quantitatively evaluate the flow estimation error $||F_M(t) - F_L(t)||$ as
\begin{gather}
    p_E = \frac{|| F_M(t) - F_L(t)||}{2 max (\hat{F}_{Lmax}, \hat{F}_{Mmax})}
\label{eq 14}
\end{gather},
where $\hat{F}_{Lmax} = max(||F_L(\tau)||_{\tau \in [0,t]})$ is the maximum algorithm-estimated flow speed until time $t$, and $\hat{F}_{Mmax} = max(||F_M(\tau)||_{\tau \in [0,t]})$ is the maximum glider-estimated flow speed until time $t$. If flow estimation error is too large ($p_E > \gamma_f$, where $\gamma_f$ is a pre-selected threshold), the detected anomaly is likely a false alarm.

\section{Experimental Evaluation}
\label{experimental evalution}

We apply the anomaly detection algorithm to four glider deployments across the coastal ocean of Florida and Georgia, USA. For evaluation, the anomaly detected by the algorithm is cross-validated by high-resolution glider DBD data and pilot notes. In particular, we simulate the online detection process on SBD data and compare the result with that detected from DBD data. For reference, the designed parameters are listed in TABLE~\ref{table parameters}.

\begin{table*}[t]
\caption{Designed parameters of experiments}
\begin{center}
\renewcommand{\arraystretch}{1.5}
\begin{tabular}{| c || c | c |  c | c | c |}
 \hline
 Parameters & Franklin & USF-Sam  &  USF-Gansett & USF-Stella   \\
 \hline
 
 \hline
  number of basis functions N  & 4 & 4 & 4 & 4   \\
  
 \hline
  width $\sigma_i$ & 13e3 & 50e3 & 32e3  & 30e3      \\
  
 \hline
  tidal phase $\upsilon_i$ & 0 & 0 &  0  &   0    \\
  
   \hline
  tidal frequency $\omega_i$  & $2\pi$e-6   & $2\pi$e-6 & $2\pi$e-6   & $2\pi$e-6   \\
  
 \hline 
  gain $K$  &   $\begin{bmatrix} 0.003 & 0 \\  0 & 0.003 \end{bmatrix}$ &   $\begin{bmatrix} 0.002 & 0 \\  0 & 0.002 \end{bmatrix}$    &   $\begin{bmatrix} 0.003 & 0 \\  0 & 0.003 \end{bmatrix}$  &  $\begin{bmatrix} 0.003 & 0 \\  0 & 0.003 \end{bmatrix}$  \\
   
  \hline  
  gain $\Bar{\gamma}$  & 5e-7  &  5e-7& 1e-6 & 1e-7  \\
  
  \hline
  gain $ s $ & 30e-3  & 7e-3  & 18e-3  & 30e-3    \\
  
  \hline
  false alarm threshold $\gamma_f$ & 1.0  & 1.0  & 1.0 & 1.0  \\

  \hline 
  maximum glider speed $V_{max}$  & 0.25 & 0.25  & 0.25  & 0.25   \\

  \hline
  minimum glider speed  $V_{min}$  & 0.15  & 0.15  & 0.15  & 0.15  \\

  \hline
\end{tabular}
\label{table parameters}
\end{center}
\end{table*}

\subsection{Experimental Setup}

All the gliders used in the experiments are Slocum gliders, which are a type of autonomous underwater vehicles (AUVs) that move by adjusting buoyancy and center of gravity \cite{robustandready}.  These gliders are able to perform ocean surveying for months by traveling at $0.25–0.35 m \cdot s^{-1}$. During the mission, the gliders surface at fixed intervals (usually four hours) to transmit lightweight SBD datasets to the on-shore dockserver. It also estimates the average flow speed through dead reckoning. Post recovery, all the datasets are downloaded off the glider and stored as DBD files. The anomaly detection algorithm in Section \ref{anomaly detection algo} is applied to post-recovery DBD data and real-time SBD data in the offline and online mode, respectively, and verified by both the sensor data segments and the glider pilots' notes.

\begin{figure}[ht]
     \centering

     \hfill
     \begin{subfigure}[b]{0.4\textwidth}
         \centerline{\includegraphics[width=\textwidth, height=3cm]{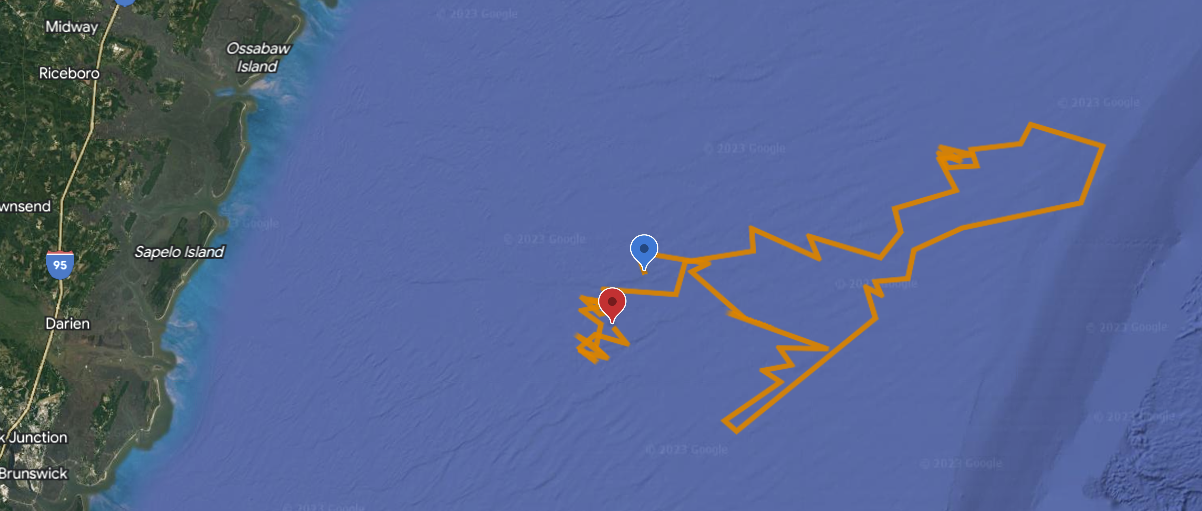}}
          \caption{Google Earth trajectory for the 2022 Franklin deployment.}
         \label{franklin google earth trajectory}
     \end{subfigure}
     
     \hfill
     \begin{subfigure}[b]{0.4\textwidth}
         \centerline{\includegraphics[width=\textwidth, height=3cm]{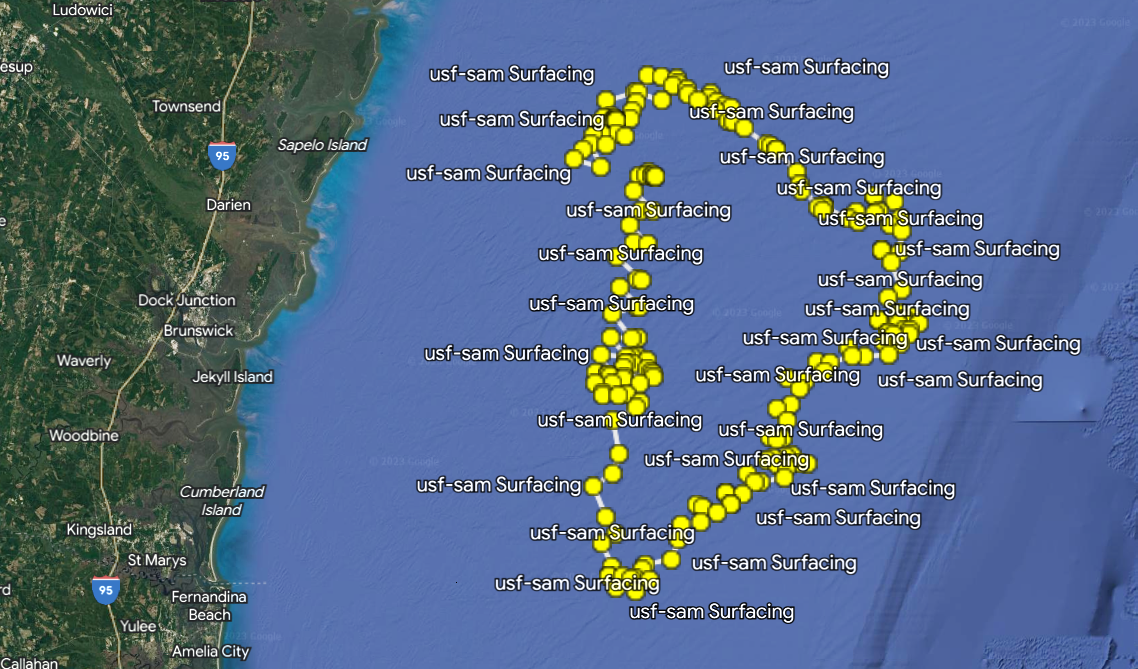}}
         \caption{Google Earth trajectory for the 2023 USF-Sam deployment.}
         \label{usf-sam google earth trajectory}
     \end{subfigure}

     \hfill
     \begin{subfigure}[b]{0.4\textwidth}
         \centerline{\includegraphics[width=\textwidth, height=3cm]{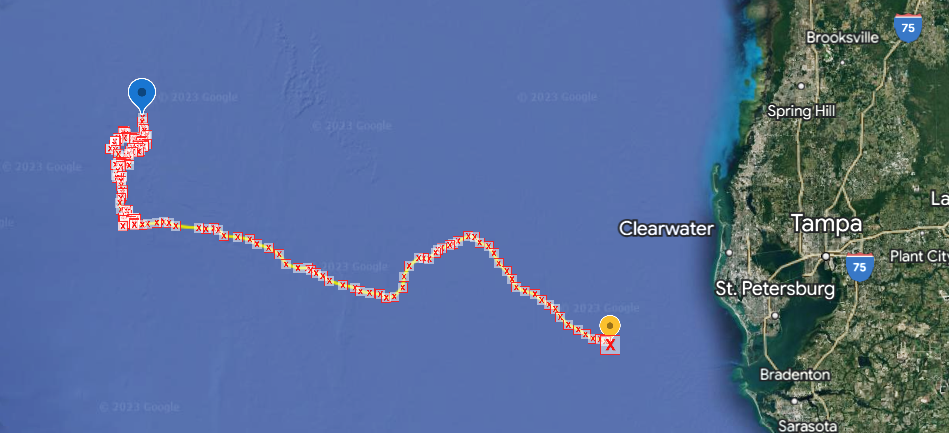}}
         \caption{Google Earth trajectory for the 2021 USF-Gansett deployment.}
         \label{usf-gansett google earth trajectory}
     \end{subfigure}

    \hfill
     \begin{subfigure}[b]{0.4\textwidth}
         \centerline{\includegraphics[width=\textwidth, height=3cm]{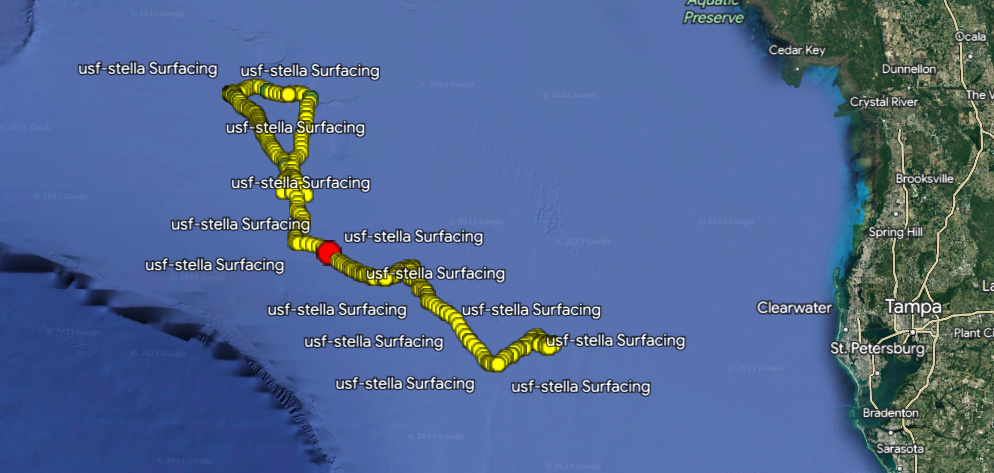}}
        \caption{Google Earth trajectory for the 2023 USF-Stella deployment.}
      \label{usf-stella google earth trajectory}
     \end{subfigure}
     
        \caption{Google Earth trajectories of glider deployments}
        \label{google earth trajectories}
\end{figure}

The deployment details of four glider deployments are shown in TABLE~\ref{table deployments} along with the Google Earth trajectories in Fig.~\ref{google earth trajectories}. It is worth mentioning that USF-Sam is piloted under the support of GENIoS\_Python \cite{yang2023real} in real time, and USF-Sam is simulated by the online detection algorithm to report any potential anomaly.

\begin{table*}[t]
\caption{Deployment details}
\begin{center}
\renewcommand{\arraystretch}{1.5}
\begin{tabular}{| c || c | c |  c | c | c |}
  \hline
  Deployments & Time (UTC) & Area  &  Mission distance & Glider team & Algorithm  \\
  \hline
 
  \hline
  Franklin   & Oct. 08 - Nov. 01, 2022 & Savannah, Georgia &  600 km & SkIO & Anomaly Detection   \\
  
  \hline
  USF-Sam & Feb. 25 - Mar. 27, 2023  & Gray's Reef  & 610 km  & USF   &  Anomaly Detection \&  GENIoS\_Python  \\
  
  \hline
  USF-Gansett  & Nov. 10 - Dec. 8, 2021 & Tampa Bay, Florida &  1100 km  &   USF  &  Anomaly Detection \\
  
  \hline
  USF-Stella  &   Mar. 24 - May 09, 2023   &  Clearwater, Florida  & 400 km   & USF &  Anomaly Detection\\
 
  \hline
\end{tabular}
\label{table deployments}
\end{center}
\end{table*}

\subsection{Large-scale Experiments}

The large-scale experiments apply hindcast anomaly detection to full resolution DBD files downloaded from the glider on the shore. For verification, the algorithm-detected anomaly is compared with that directly seen from post-mission DBD data with the highest possible resolution and pilot logs.

\subsubsection{Franklin}
\label{franklin}

\begin{figure}[ht]
     \centering

     \hfill
     \begin{subfigure}[b]{0.48\textwidth}
         \centerline{\includegraphics[width=\textwidth, height=3cm]{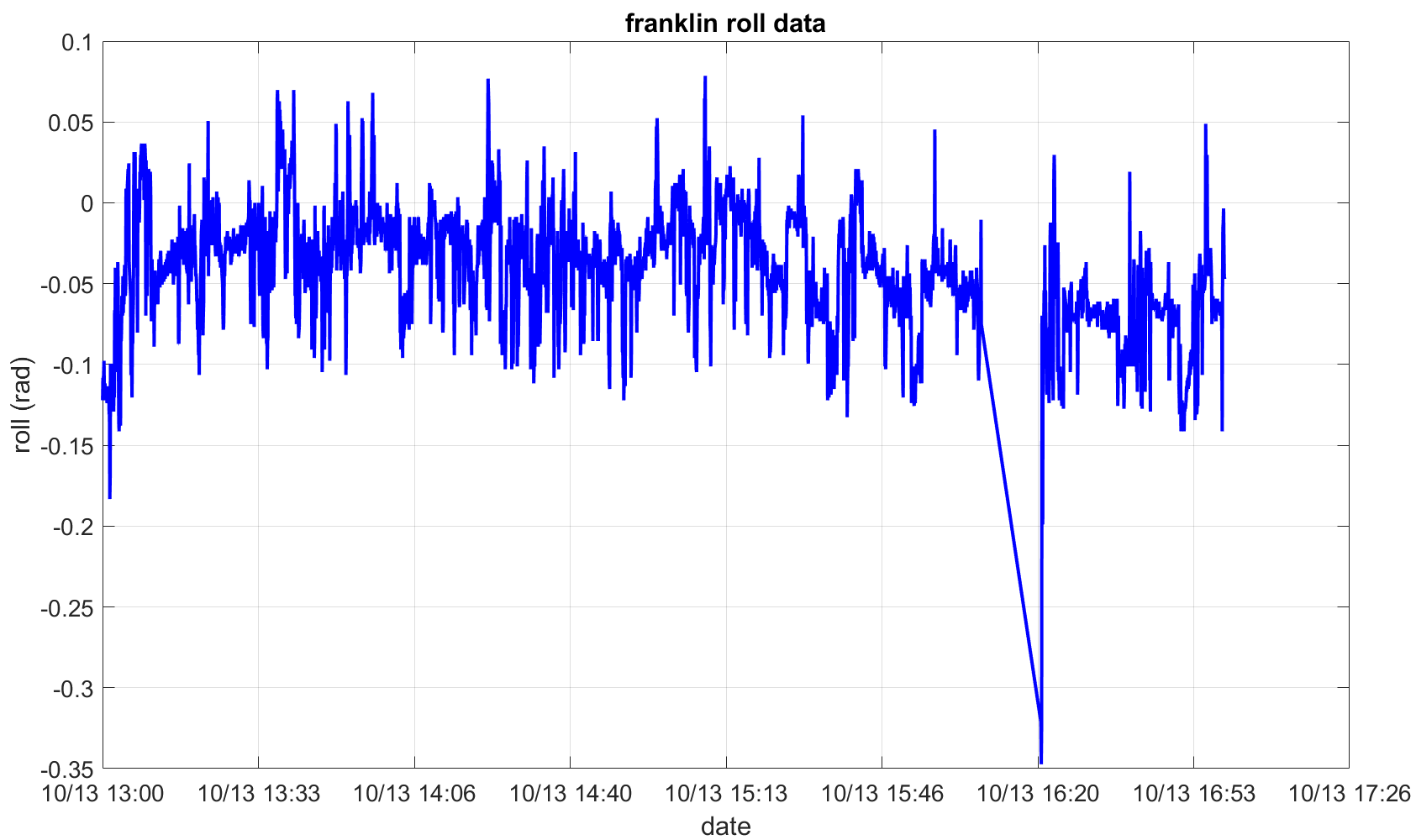}}
         \caption{Glider-measured roll (rad) from post-recovery DBD data.}
         \label{franklin roll}
     \end{subfigure}
       
     \hfill
     \begin{subfigure}[b]{0.48\textwidth}
         \centerline{\includegraphics[width=\textwidth, height=3cm]{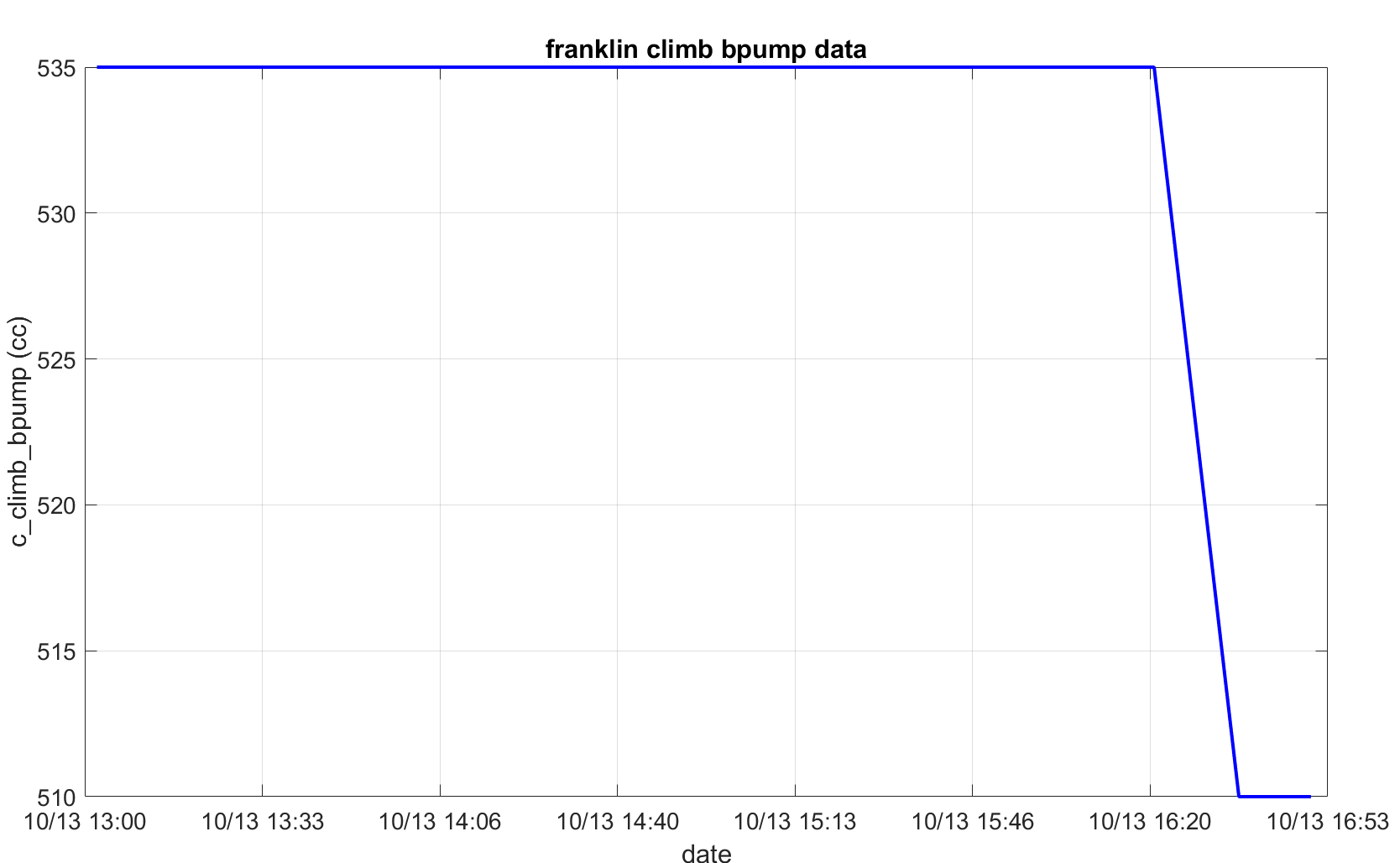}}
         \caption{Glider-measured climb ballast (cc) from post-recovery DBD data.}
         \label{franklin climb}
     \end{subfigure}
     
        \caption{ground truth for the 2022 Franklin deployment.}
        \label{franklin ground truth data}
    \end{figure}

From October 12 to October 13, 2022, Franklin experienced two aborts and delays of up to 40 minutes of subsequent surfacings. The glider pilot time believed that Franklin had attracted remoras or had encountered an obstruction on his port wing, resulting in a roll change shown in Fig.~\ref{franklin roll}. Its climb to the ocean surface is unexpectedly slow even though flying with climb ballast near the upper limits of extended buoyancy pump, as shown in Fig.~\ref{franklin climb}.

\begin{figure}[ht]
        \centerline{\includegraphics[width=0.48\textwidth]{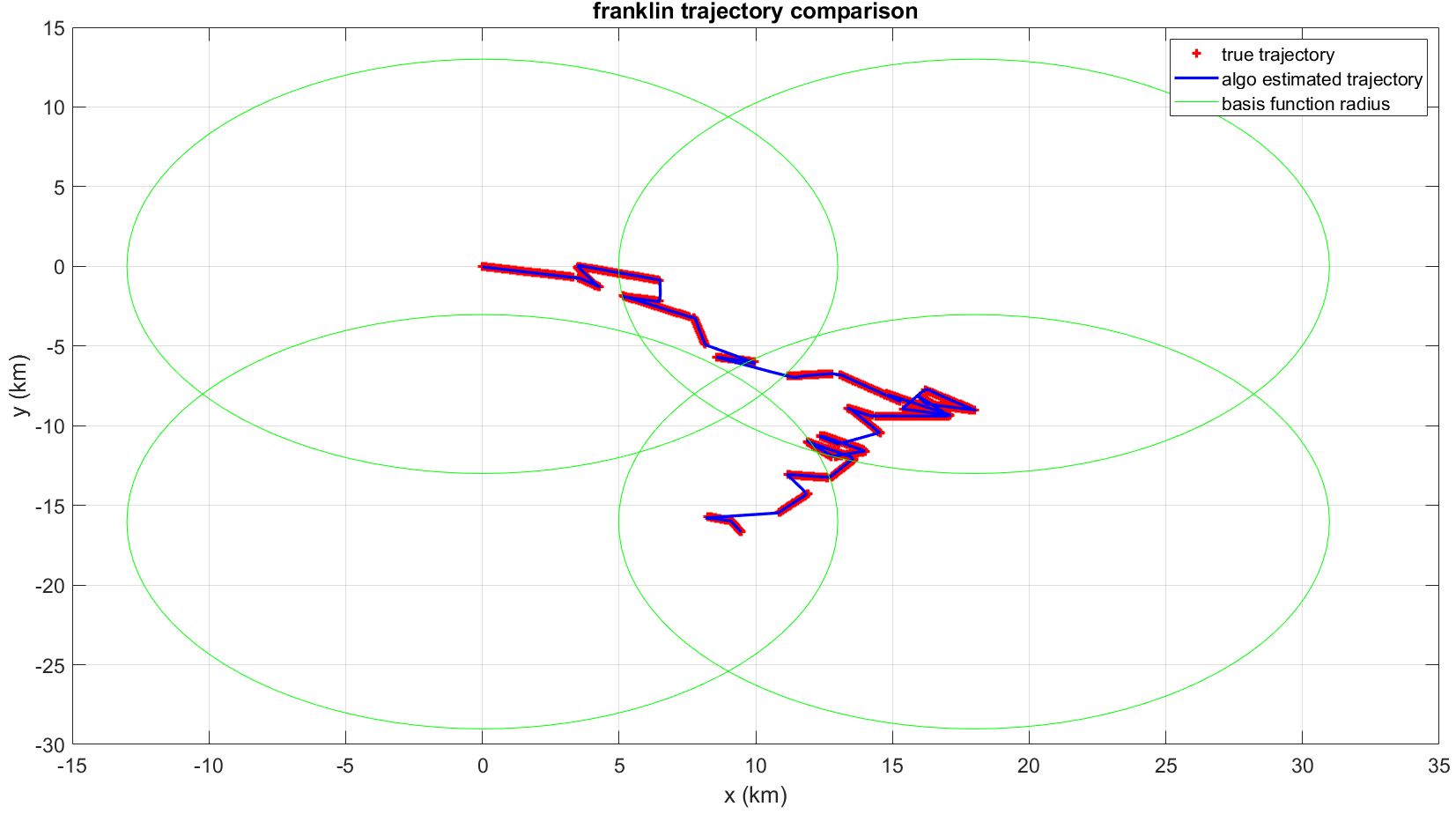}}
          \caption{Comparison of the estimated (blue) and true (red) trajectory for the 2022 Franklin deployment. The four green circles are the four basis functions covering the whole trajectory.}
        \label{franklin trajectory comparison}
    \end{figure}

\begin{figure}[ht]
        \centerline{\includegraphics[width=0.48\textwidth]{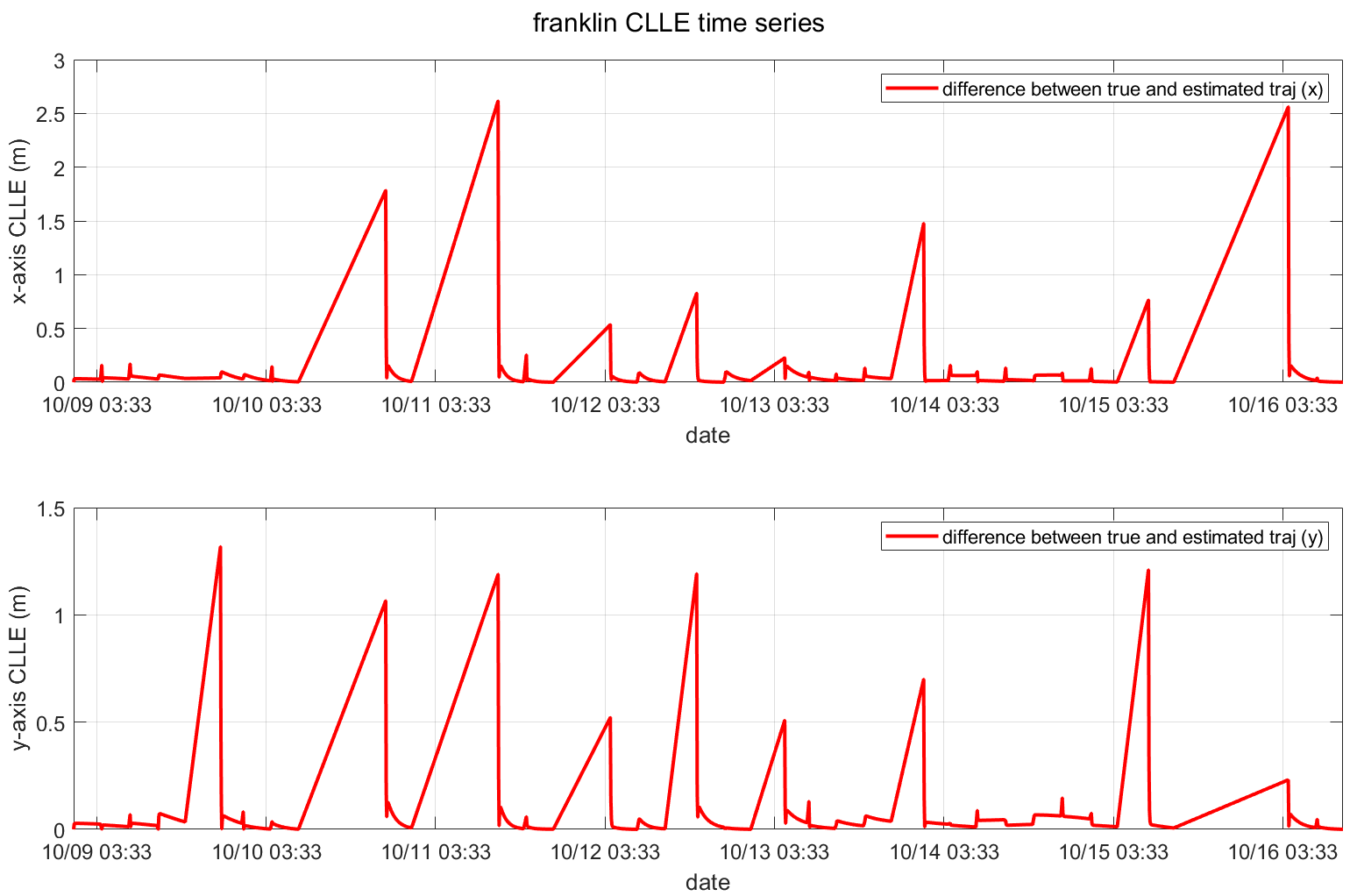}}
        \caption{CLLE (m) for the 2022 Franklin deployment.}
        \label{franklin CLLE}
    \end{figure}

Applied to the DBD data downloaded from the glider, the detection algorithm guarantees convergence of the estimated trajectory to the true trajectory as shown in Fig.~\ref{franklin trajectory comparison}. There are four basis functions (four green circles) covering the glider trajectory in the whole flow fields, which is an essential condition for parameter estimation to converge. As shown in Fig.~\ref{franklin CLLE}, the maximum CLLE is small enough as $2.5m$, considering the glider moves hundreds of kilometers in the entire deployment, so we can also conclude the convergence of CLLE.

\begin{figure}[ht]
     \centering

     \hfill
     \begin{subfigure}[b]{0.48\textwidth}
         \centerline{\includegraphics[width=\textwidth, height=3.5cm]{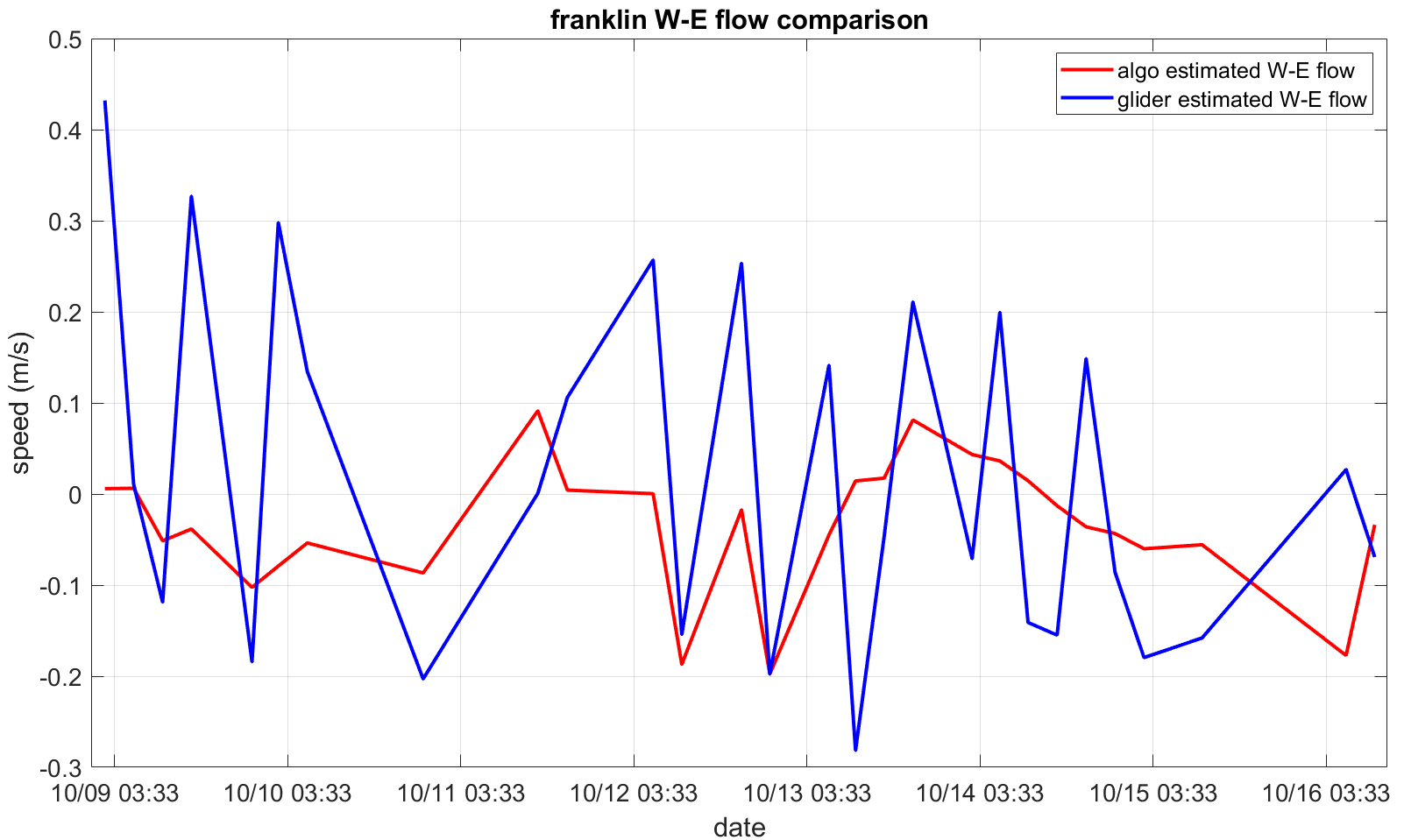}}
           \caption{W-E flow component.}
         \label{franklin W-E flow}
     \end{subfigure}

     \hfill
     \begin{subfigure}[b]{0.48\textwidth}
         \centerline{\includegraphics[width=\textwidth, height=3.5cm]{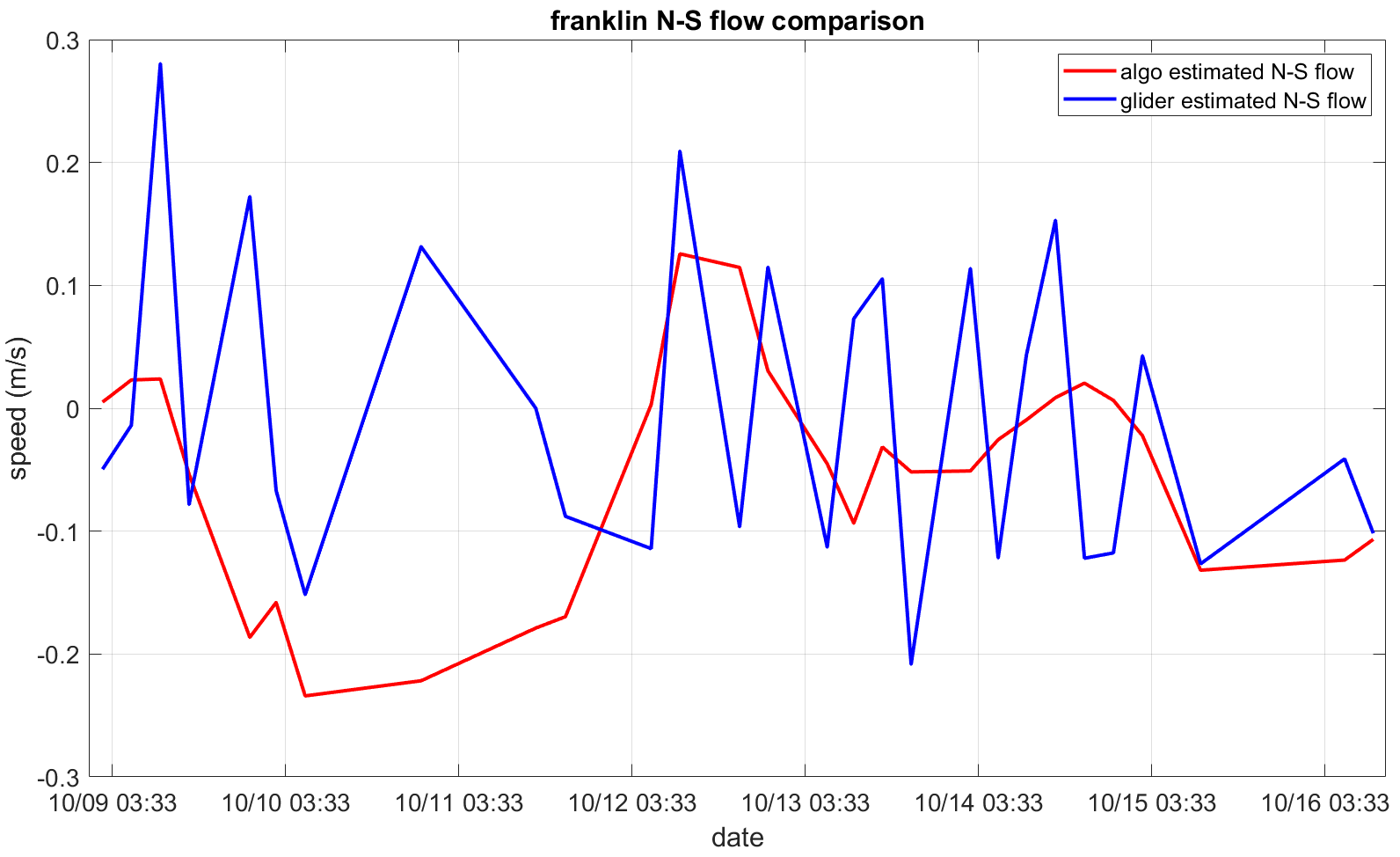}}  
           \caption{N-S flow component.}
         \label{franklin N-S flow}
     \end{subfigure}
     
      \caption{Comparison of glider-estimated and algorithm-estimated W-E ($u$, upper) and N-S ($v$, lower) flow velocities for the 2022 Franklin deployment.}
        \label{franklin flow comparison}
    \end{figure}

\begin{figure}[ht]
        \centerline{\includegraphics[width=0.48\textwidth, height=4.5cm]{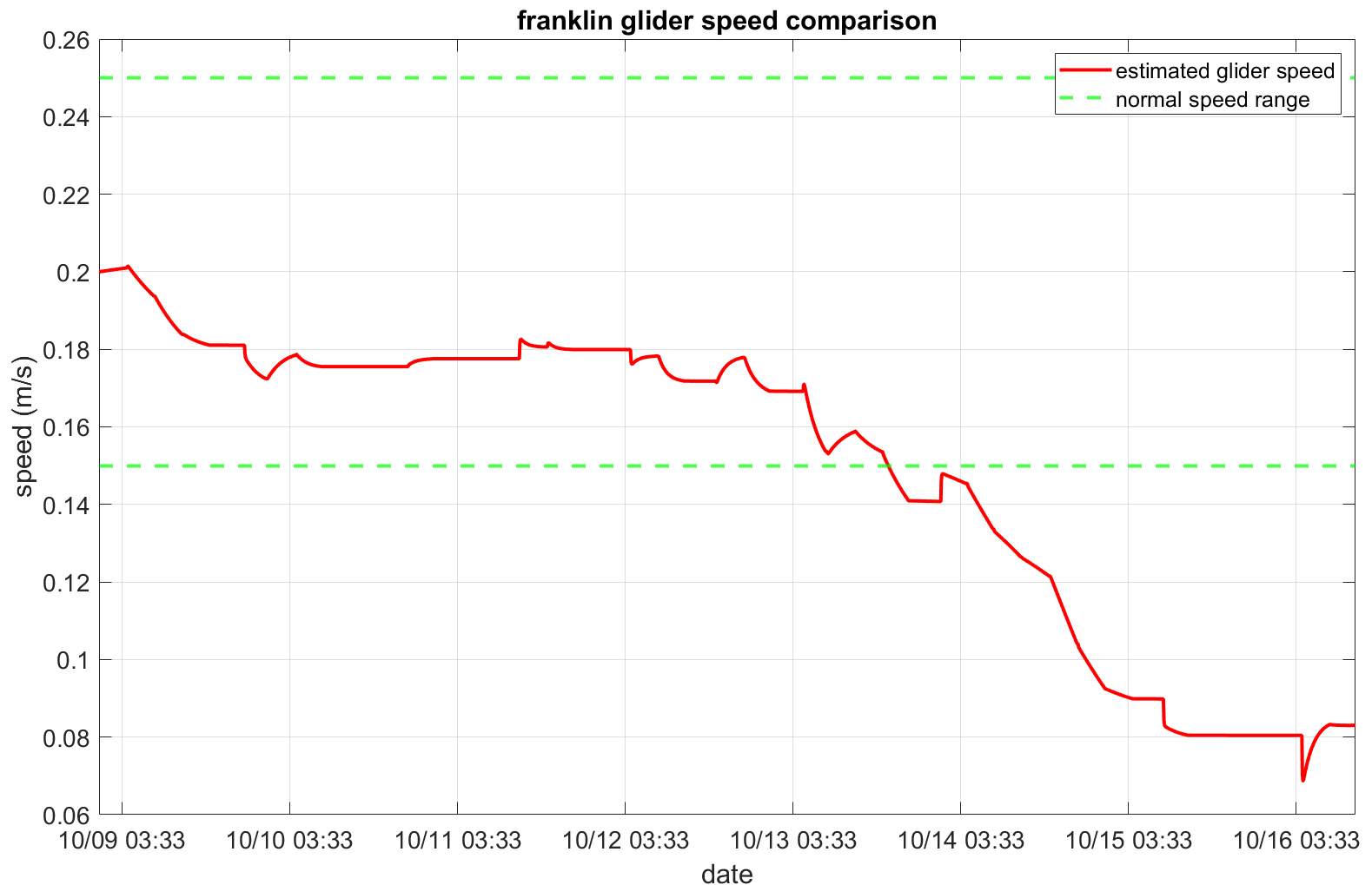}}
        \caption{Comparison of estimated glider speed (red) and normal speed range (green) for the 2022 Franklin deployment.}
        \label{franklin speed comparison}
    \end{figure}

The CLLE convergence guarantees the convergence of both the glider speed estimate and the flow speed estimate. For precise comparison, the flow is divided into its West-East (W-E or zonal) component, denoted as $u$, and  its North-South (N-S or meridional) component, denoted as $v$. The graphs in Fig.~\ref{franklin W-E flow} demonstrate that the algorithm-estimated W-E flow is close to the corresponding glider estimate, indicating minimal error in the $u$ flow estimation. A similar comparison can be observed for the N-S flow, as shown in Fig.~\ref{franklin N-S flow}. This comparative analysis provides reliability of the anomaly detection when it is triggered. If the estimated glider speed drops out of the normal speed range, the anomaly should have occurred. As shown in Fig.~\ref{franklin speed comparison},  the estimated glider speed drops out of the normal speed range (green dot line) at around  October 13, 2022, 15:00 UTC. The timestamp when the anomaly is detected by the algorithm corresponds to the timestamp detected from the glider team's report and the post-recovery DBD data. Therefore, the algorithm is verified by successfully detecting the anomaly.

\subsubsection{USF-Sam}
\begin{figure}[ht]
     \centering

     \hfill
     \begin{subfigure}[b]{0.48\textwidth}
         \centerline{\includegraphics[width=\textwidth, height=3cm]{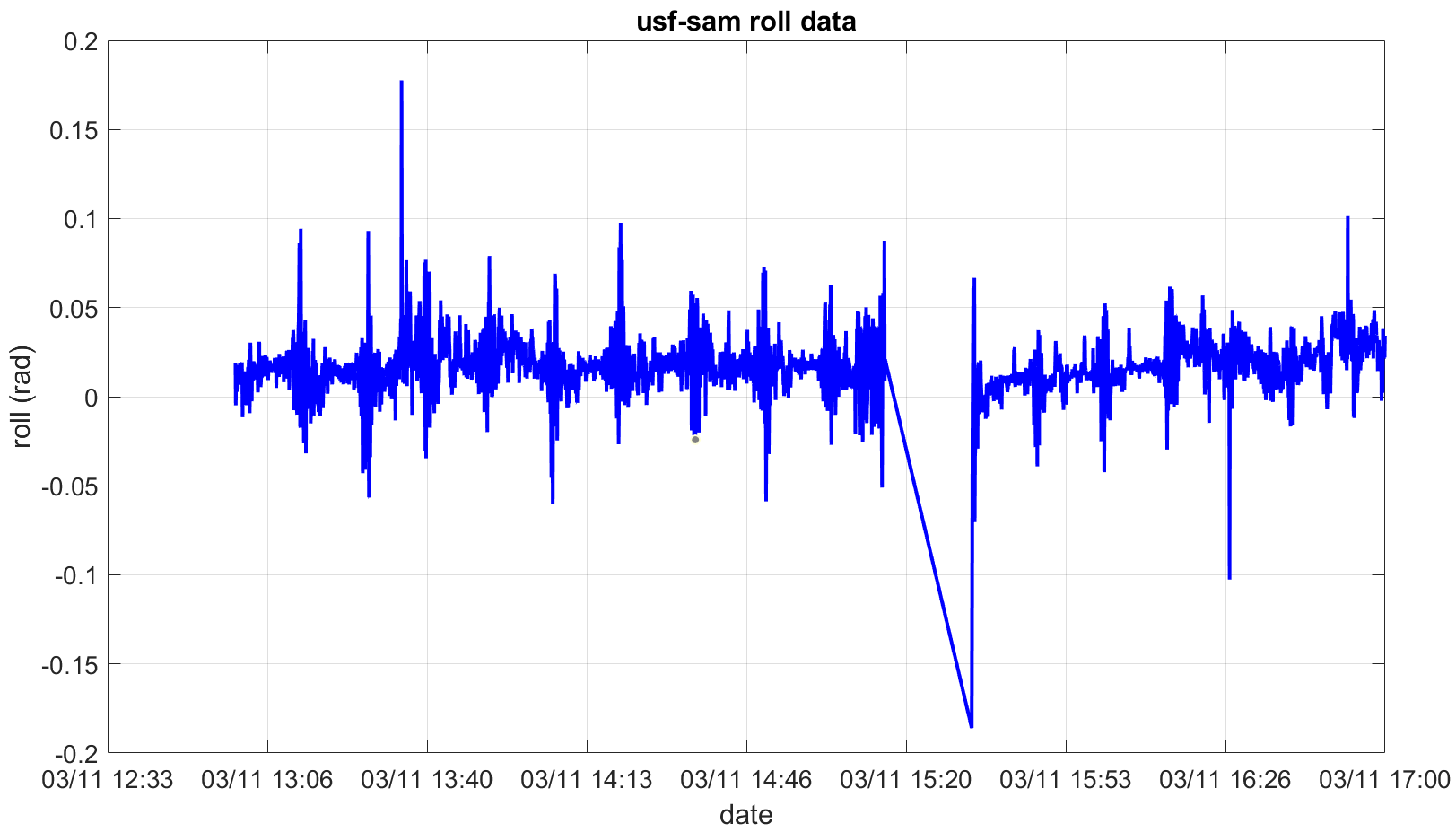}}
         \caption{Glider-measured roll (rad) from post-recovery DBD data.}
         \label{usf-sam roll}
     \end{subfigure}
     
    \hfill
     \begin{subfigure}[b]{0.48\textwidth}
         \centerline{\includegraphics[width=\textwidth, height=3cm]{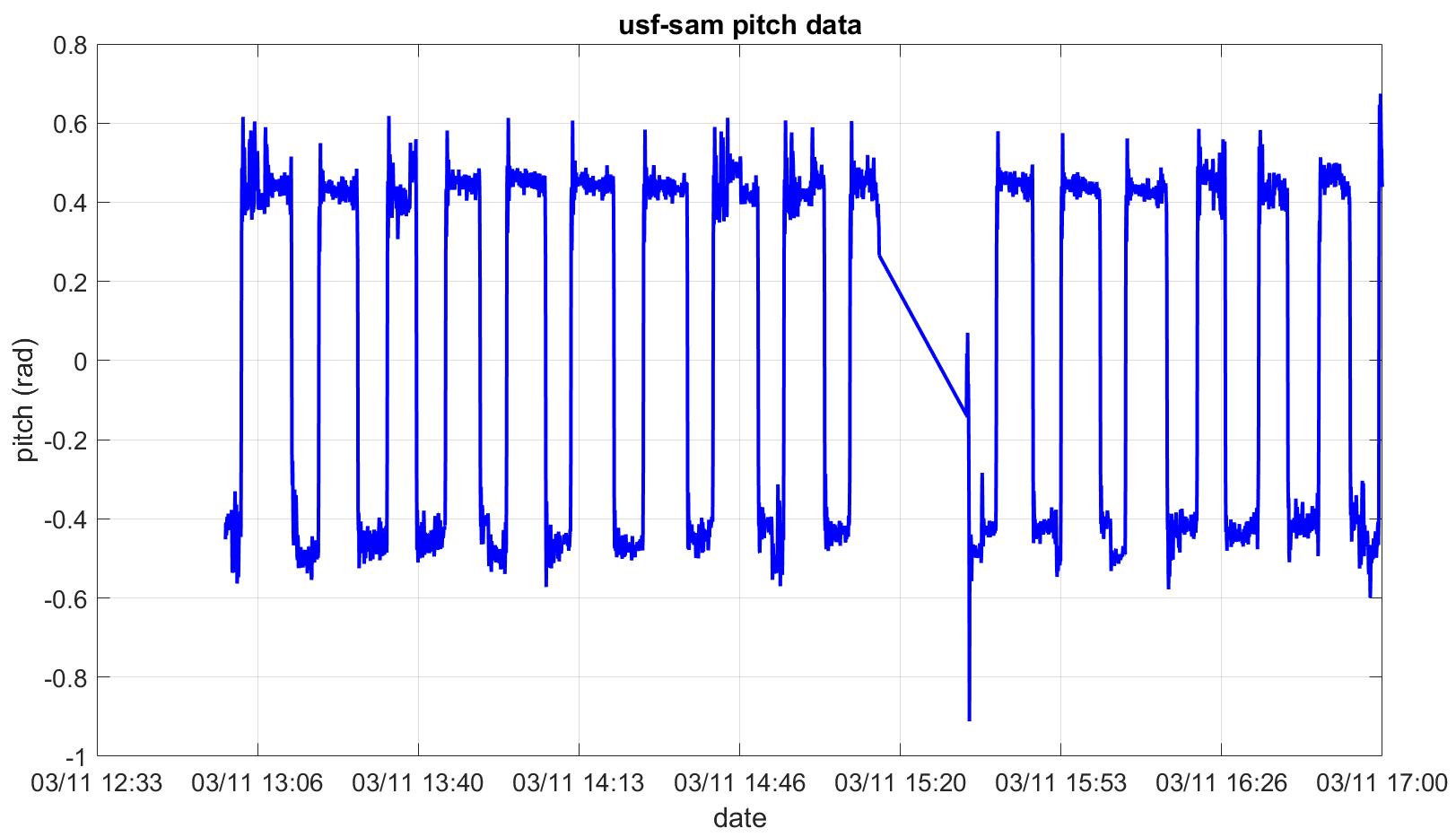}}
         \caption{Glider-measured pitch (rad) from post-recovery DBD data.}
         \label{usf-sam pitch}
     \end{subfigure}

        \caption{ground truth for the 2023 USF-Sam deployment.}
        \label{usf-sam ground truth data}
    \end{figure}

During the mission, glider pilots suggest that the remora attachment should occur between March 11 and March 12, 2023 UTC when USF-Sam has a couple of roll and pitch changes shown in Fig.~\ref{usf-sam roll} and Fig.~\ref{usf-sam pitch}. This suggestion is reinforced by USF-Sam's prolonged period of being stuck at a certain depth.

\begin{figure}[ht]
        \centerline{\includegraphics[width=0.48\textwidth]{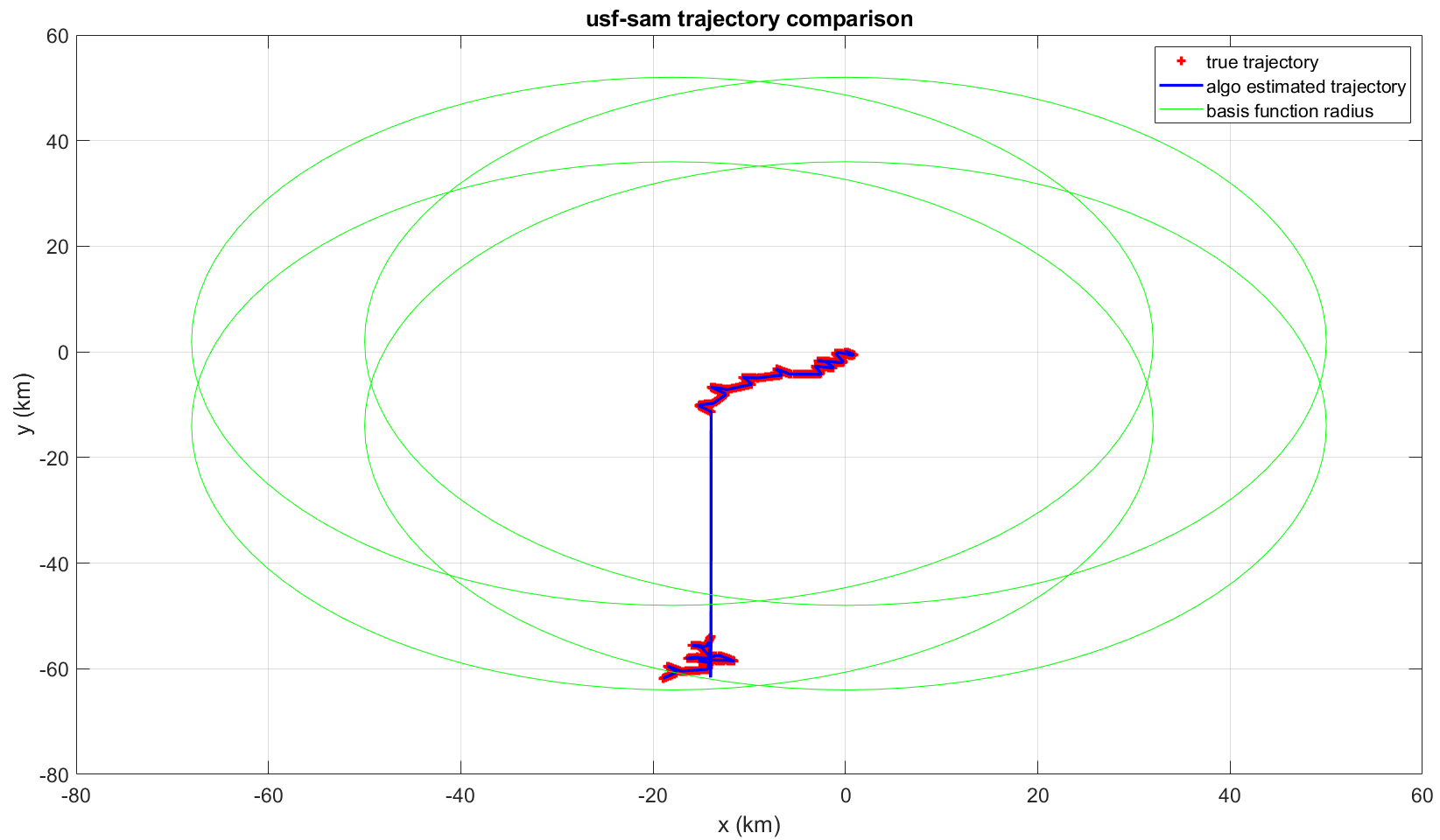}}
          \caption{Comparison of the estimated (blue) and true (red) trajectory  for the 2023 USF-Sam deployment. The four green circles are the four basis functions covering the whole trajectory.}
        \label{usf-sam trajectory comparison}
    \end{figure}

\begin{figure}[ht]
        \centerline{\includegraphics[width=0.48\textwidth]{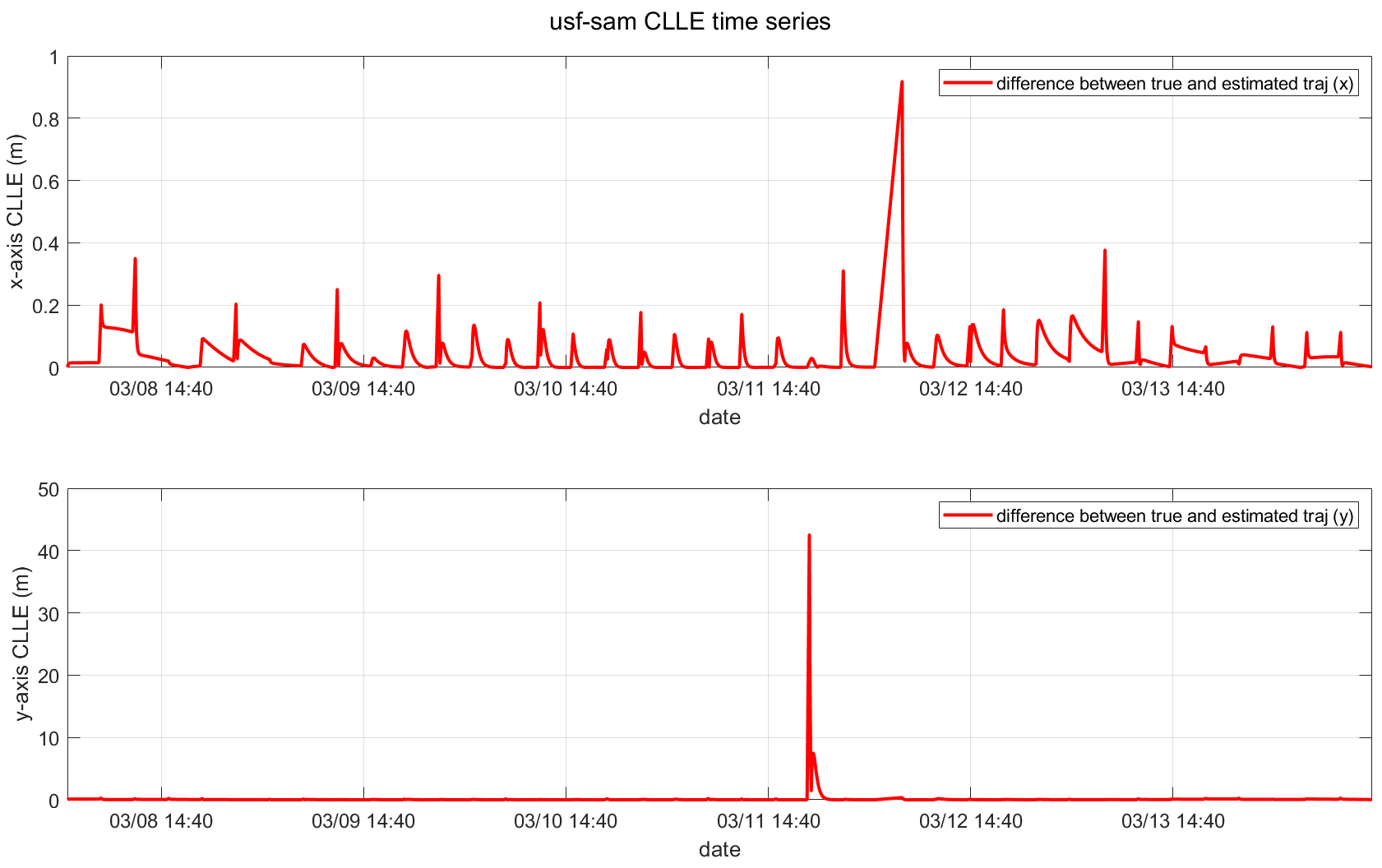}}
        \caption{CLLE (m) for the 2023 USF-Sam deployment.}
        \label{usf-sam CLLE}
    \end{figure}

\begin{figure}[ht]
     \centering

     \hfill
     \begin{subfigure}[b]{0.48\textwidth}
         \centerline{\includegraphics[width=\textwidth, height=3.5cm]{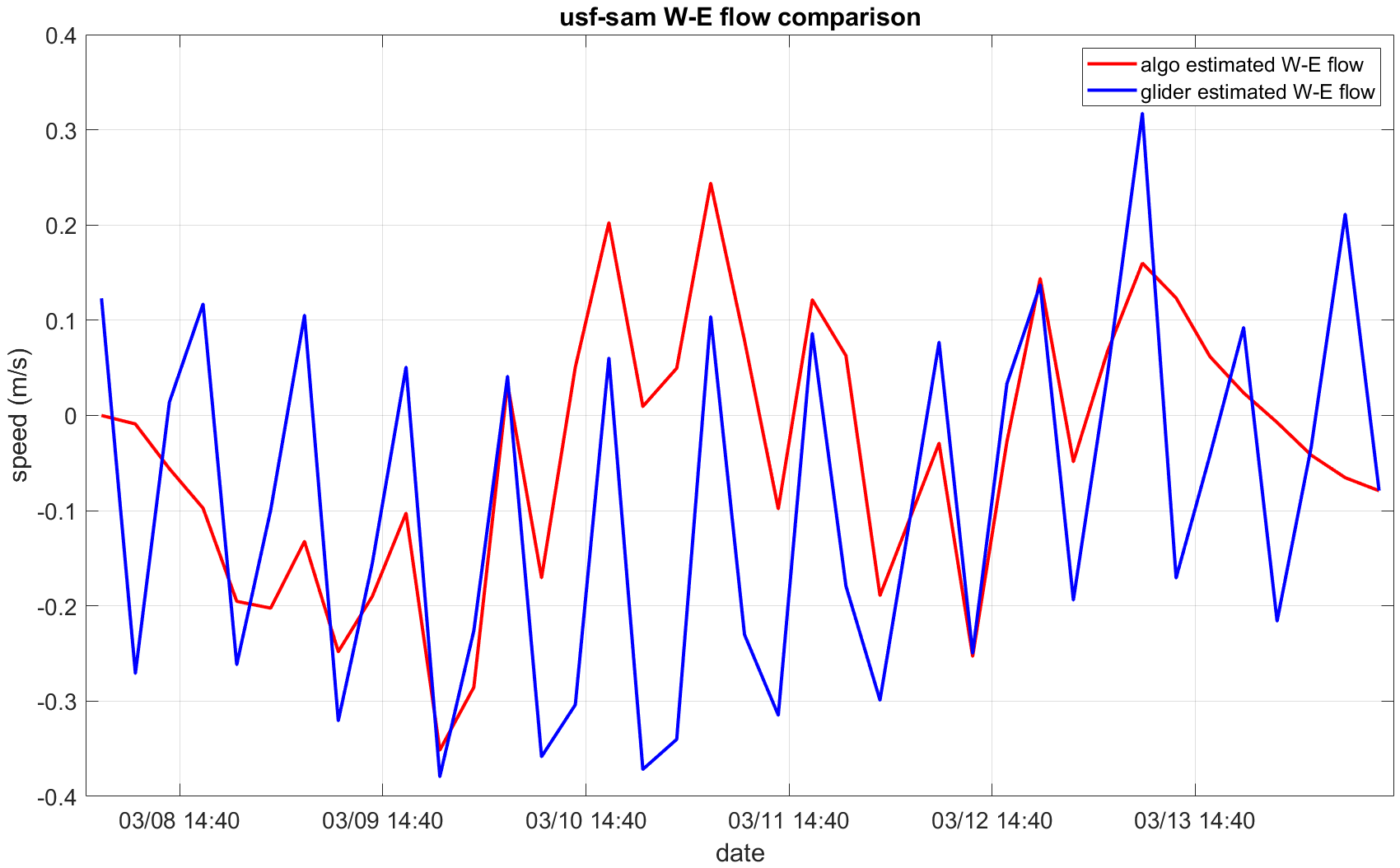}}
           \caption{W-E flow component.}
         \label{usf-sam W-E flow}
     \end{subfigure}

     \hfill
     \begin{subfigure}[b]{0.48\textwidth}
         \centerline{\includegraphics[width=\textwidth, height=3.5cm]{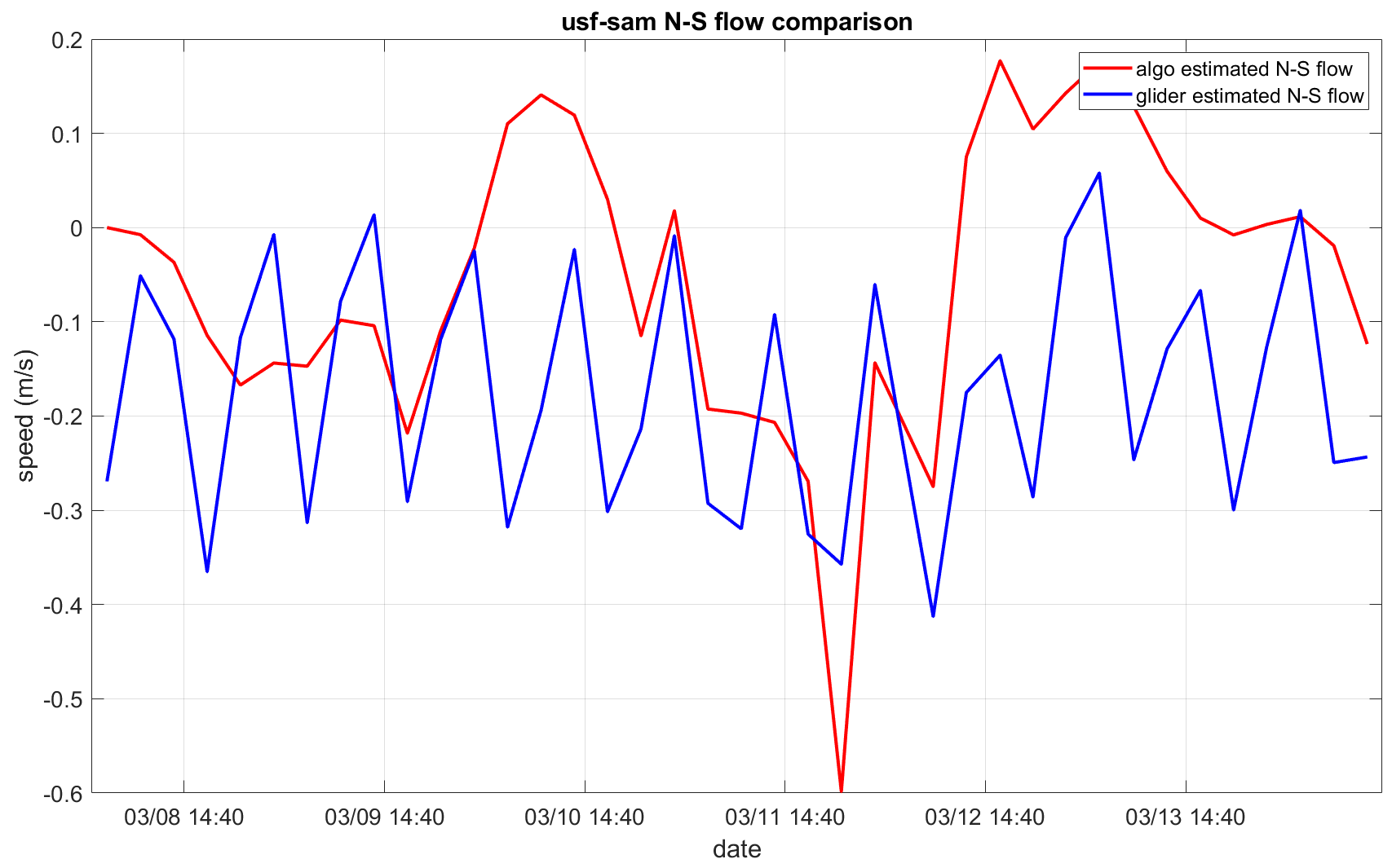}}  
           \caption{N-S flow component.}
         \label{usf-sam N-S flow}
     \end{subfigure}
     
      \caption{Comparison of glider-estimated and algorithm-estimated W-E ($u$, upper) and N-S ($v$, lower) flow velocities for the 2023 USF-Sam deployment.}
        \label{usf-sam flow comparison}
    \end{figure}

\begin{figure}[ht]
        \centerline{\includegraphics[width=0.48\textwidth]{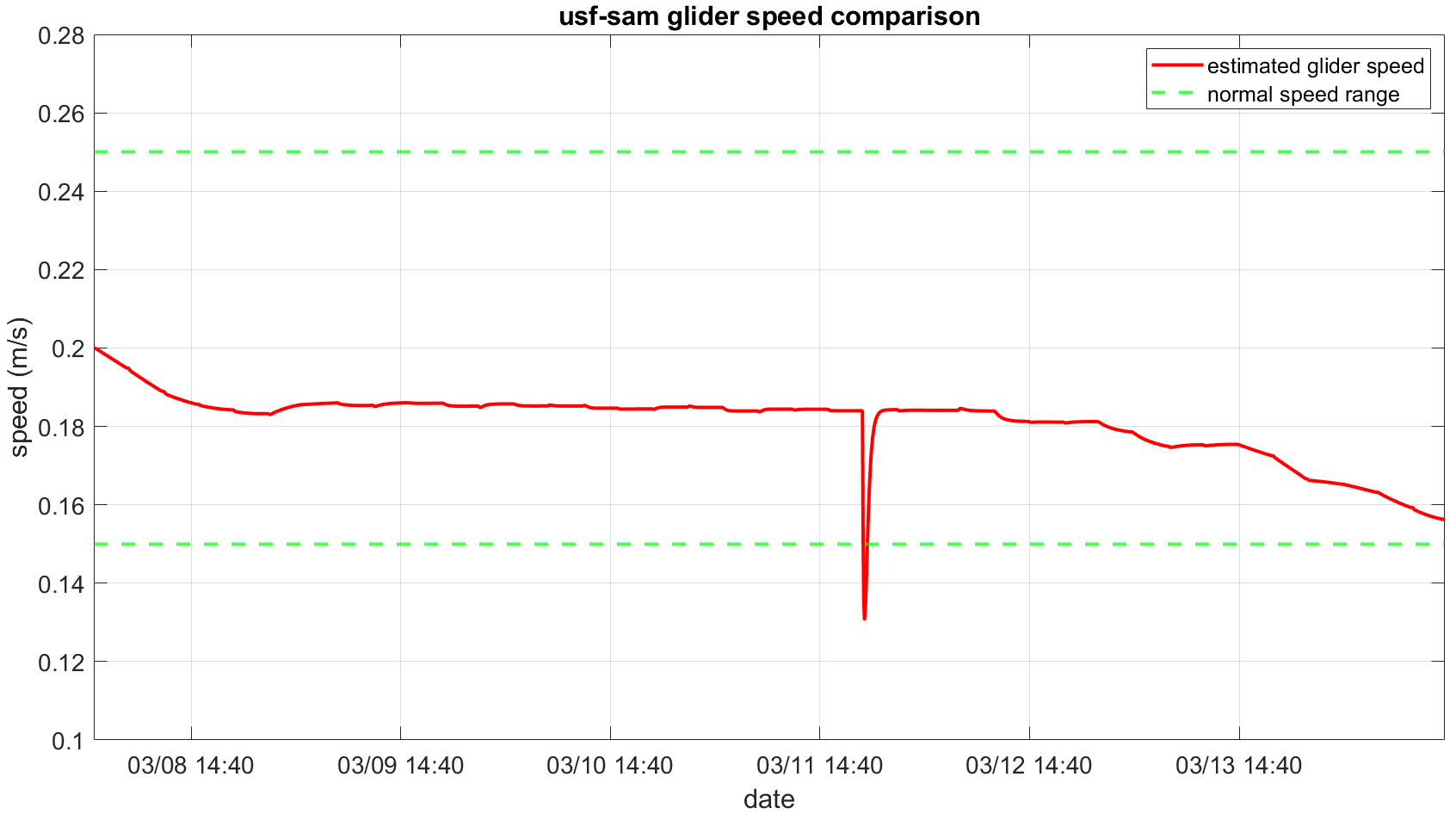}}
        \caption{Comparison of estimated glider speed (red) and normal speed range (green) for the 2023 USF-Sam deployment.}
        \label{usf-sam speed comparison}
    \end{figure}

 Based on the DBD data, the detection algorithm generates the estimated trajectory, which is close to the true trajectory as shown in Fig.~\ref{usf-sam trajectory comparison}. From quantitative analysis in Fig.~\ref{usf-sam CLLE}, the maximum CLLE $45m$ is small enough to conclude the CLLE convergence. We follow the same process of evaluating flow estimation error in Section \ref{franklin}. As shown in Fig.~\ref{usf-sam flow comparison}, the small flow estimation error suggests that the detection result can be trusted. As shown in Fig.~\ref{usf-sam speed comparison}, the estimated glider speed drops out of the normal speed range (green dot line) at around March 11, 2023, 20:00 UTC. The timestamp when the anomaly is detected by the algorithm corresponds to the timestamp detected from the glider team's report and the DBD dataset.

The simulated online experiment implements anomaly detection using subsetted real-time SBD files transmitted from the glider to the dockserver during the mission.  For example, the SBD file may contain fewer than 30 variables at 18-1800 s intervals, and is often subsampled to every 3rd or 4th yo (or down-up cycle), compared to the approximately 3000 variables stored at approximately 1 s interval on every yo in the DBD file processed on shore. The algorithm fetches new SBD files from the dockserver, parses SBD data from the SBD files, and applies the detection algorithm to the SBD data in an online mode. The online detection holds unique significance from the perspective that the detection results could help pilots monitor glider conditions in real time, thus circumventing any further loss and DBD data anomaly is only available post recovery.

\begin{figure}[ht]
        \centerline{\includegraphics[width=0.48\textwidth]{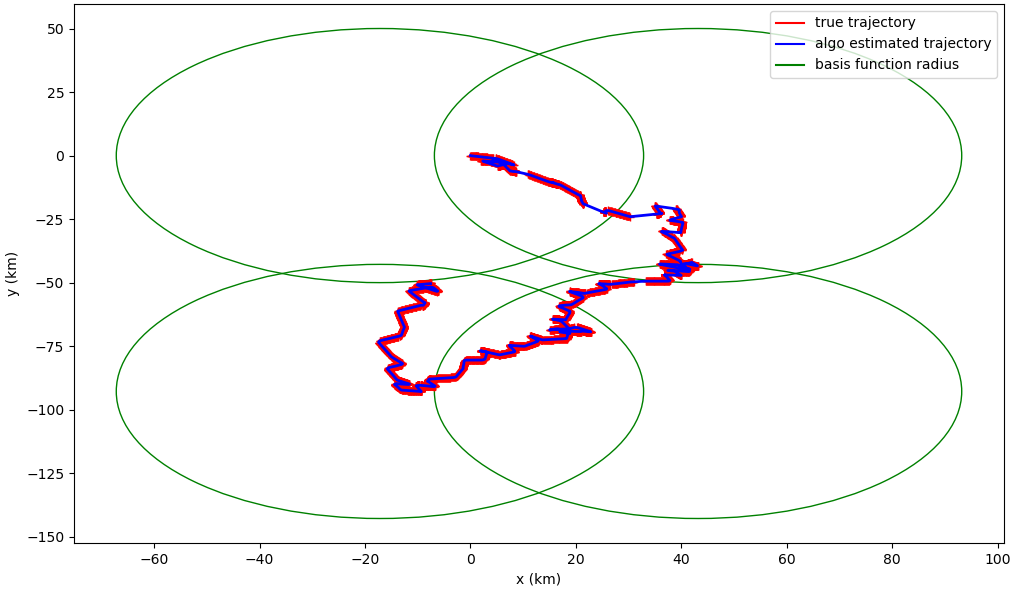}}
          \caption{Comparison of the estimated (blue) and true (red) trajectory  for the 2023 USF-Sam deployment based on real-time SBD data. The four green circles are the four basis functions covering the whole trajectory.}
        \label{usf-sam online trajectory comparison}
    \end{figure}

\begin{figure}[ht]
        \centerline{\includegraphics[width=0.48\textwidth]{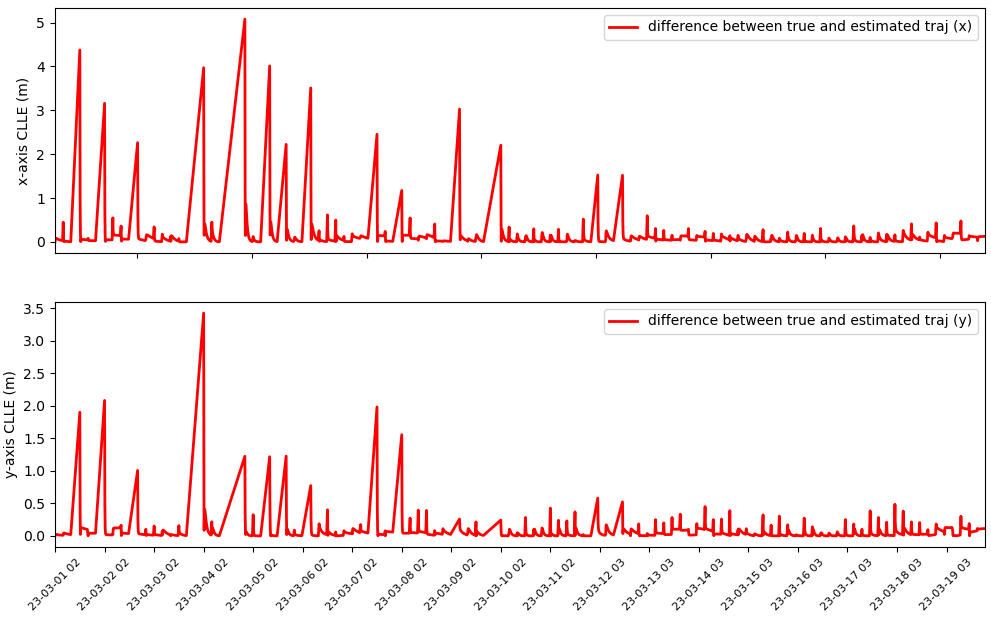}}
        \caption{CLLE (m) for the 2023 USF-Sam deployment based on real-time DBD data.}
        \label{usf-sam online CLLE}
    \end{figure}

\begin{figure}[ht]
     \centering

     \hfill
     \begin{subfigure}[b]{0.48\textwidth}
         \centerline{\includegraphics[width=\textwidth, height=3.5cm]{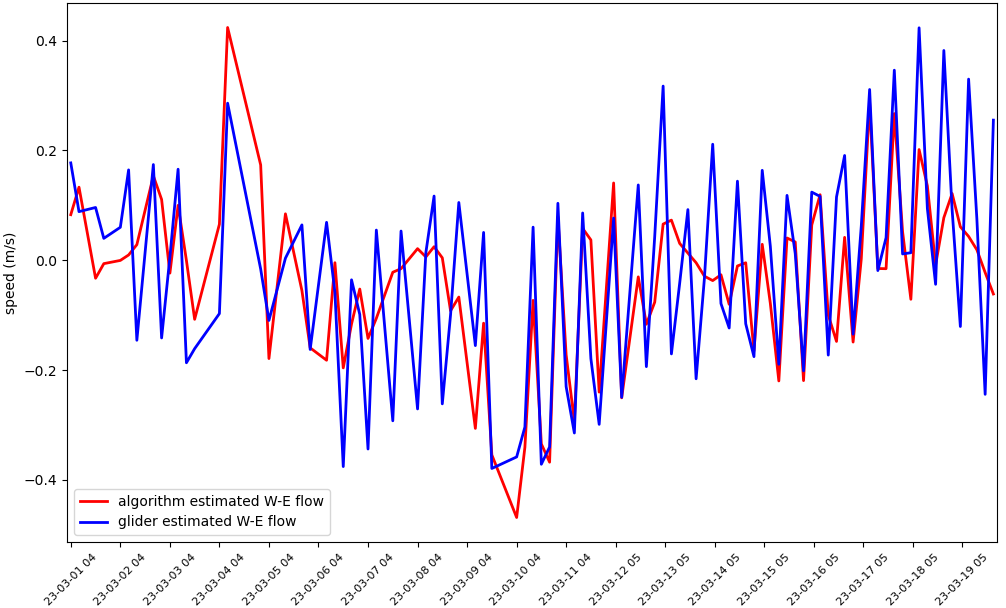}}
           \caption{W-E flow component.}
         \label{usf-sam online W-E flow}
     \end{subfigure}

     \hfill
     \begin{subfigure}[b]{0.48\textwidth}
         \centerline{\includegraphics[width=\textwidth, height=3.5cm]{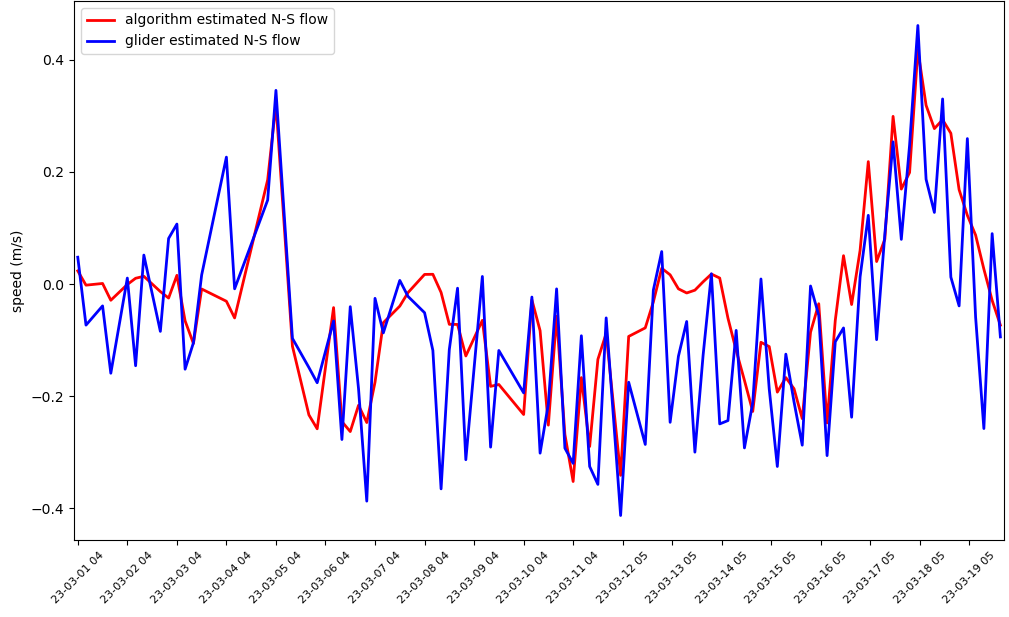}}  
           \caption{N-S flow component.}
         \label{usf-sam online N-S flow}
     \end{subfigure}
     
      \caption{Comparison of glider-estimated and algorithm-estimated W-E ($u$, upper) and N-S ($v$, lower) flow velocities for the 2023 USF-Sam deployment based on real-time SBD data.}
        \label{usf-sam online flow comparison}
    \end{figure}

\begin{figure}[ht]
        \centerline{\includegraphics[width=0.48\textwidth]{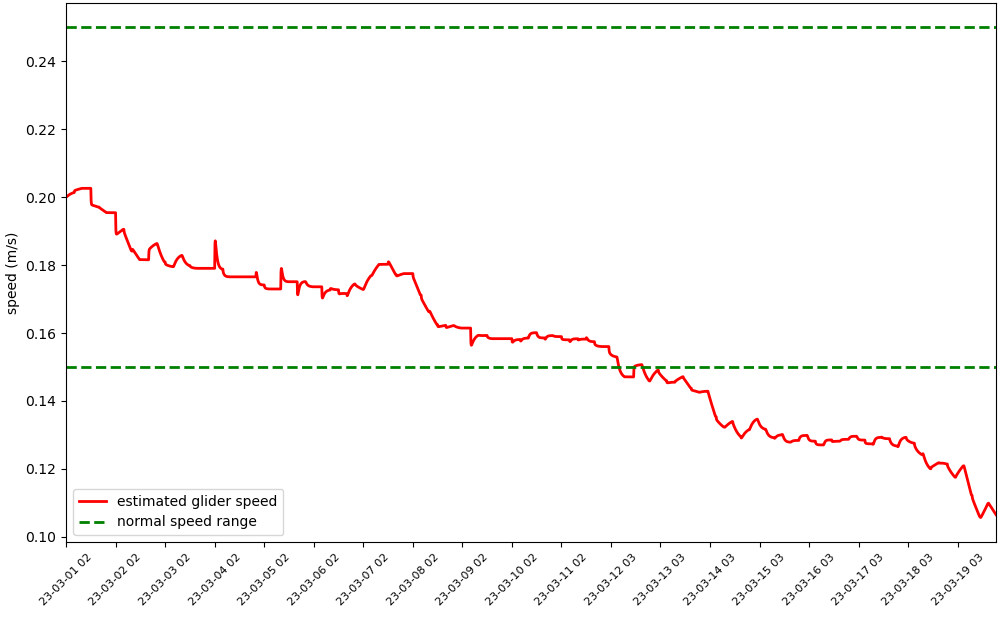}}
        \caption{Comparison of estimated glider speed (red) and normal speed range (green) for the 2023 USF-Sam deployment based on real-time SBD data.}
        \label{usf-sam online speed comparison}
    \end{figure}

Instead of waiting for the DBD data after the entire mission, the online detection is capable of utilizing the SBD data in real time. As shown in Fig.~\ref{usf-sam online trajectory comparison}, the detection algorithm can also achieve trajectory convergence similar to using the DBD data. The maximum CLLE $5m$ in Fig.~\ref{usf-sam online CLLE} is sufficiently small given the glider moving range in the ocean. Therefore, we can conclude the CLLE convergence. We follow the same process of evaluating flow estimation error in Section \ref{franklin}. As shown in Fig.~\ref{usf-sam online flow comparison}, the small flow estimation error suggests that the online detection result is trustworthy. As shown in Fig.~\ref{usf-sam online speed comparison}, the estimated glider speed drops out of the normal speed range (green dot line) at around March 12, 2023, 03:00 UTC. This result matches reasonably well the above result from the DBD data, which justifies that we can trust the online anomaly detection applied to real-time SBD data.

\subsubsection{USF-Gansett}
\begin{figure}[ht]
     \centering

     \hfill
     \begin{subfigure}[b]{0.48\textwidth}
         \centerline{\includegraphics[width=\textwidth, height=3.2cm]{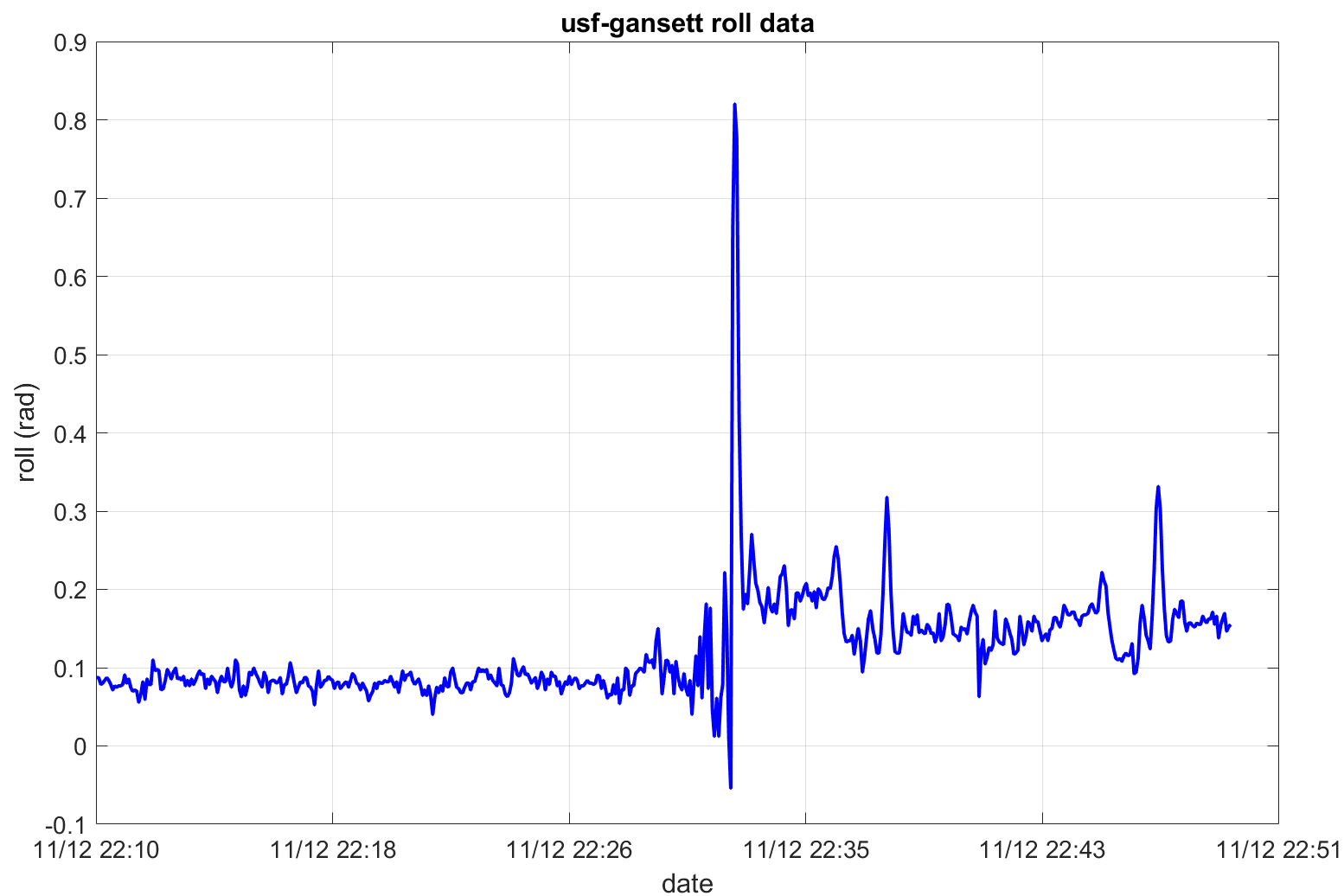}}
         \caption{Glider-measured roll (rad) from post-recovery DBD data.}
         \label{usf-gansett roll}
     \end{subfigure}

    \hfill
     \begin{subfigure}[b]{0.48\textwidth}
         \centerline{\includegraphics[width=\textwidth, height=3.2cm]{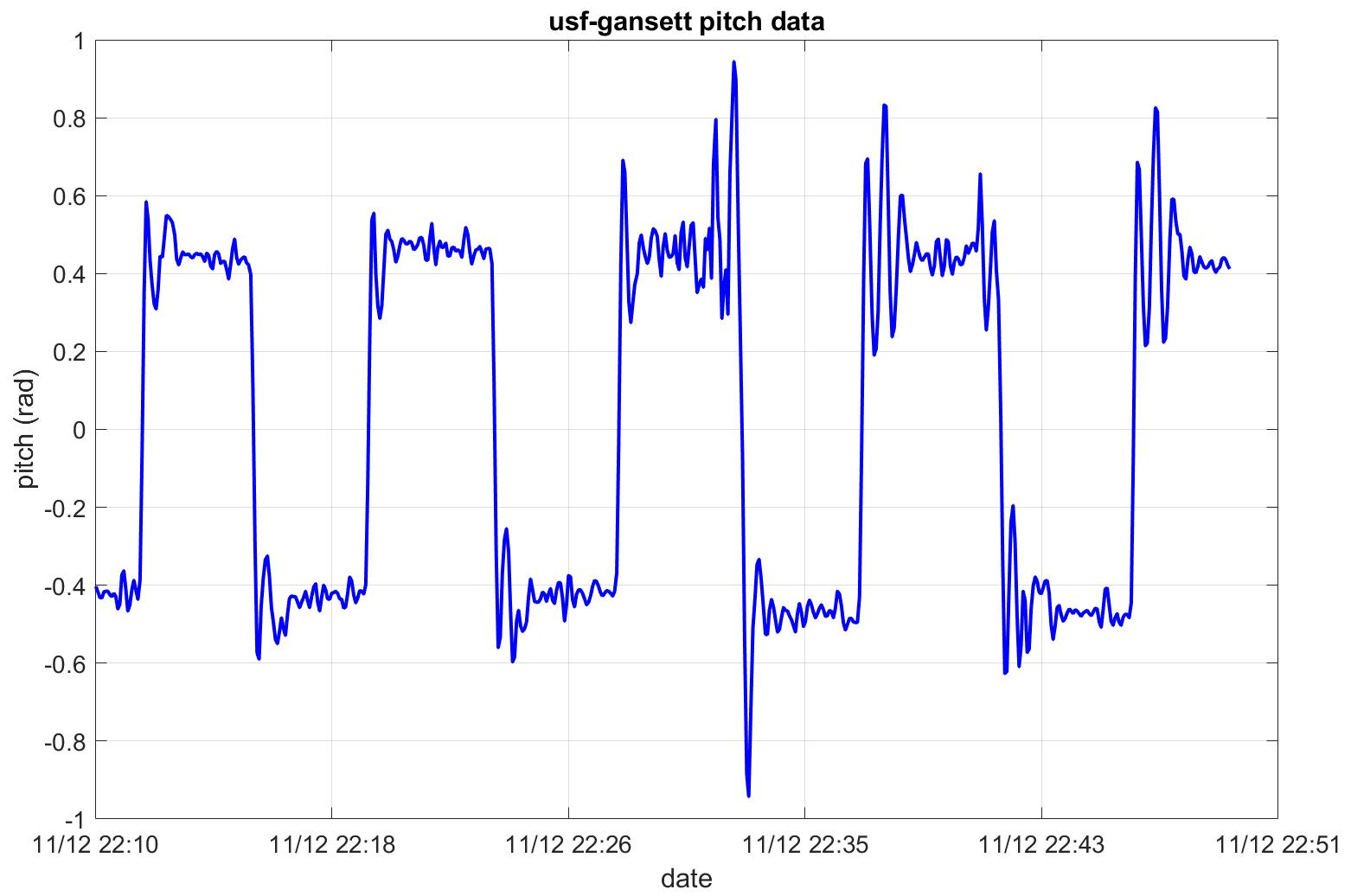}}
         \caption{Glider-measured pitch (rad) from post-recovery DBD data.}
         \label{usf-gansett pitch}
     \end{subfigure}

     \hfill
     \begin{subfigure}[b]{0.48\textwidth}
         \centerline{\includegraphics[width=\textwidth, height=3.2cm]{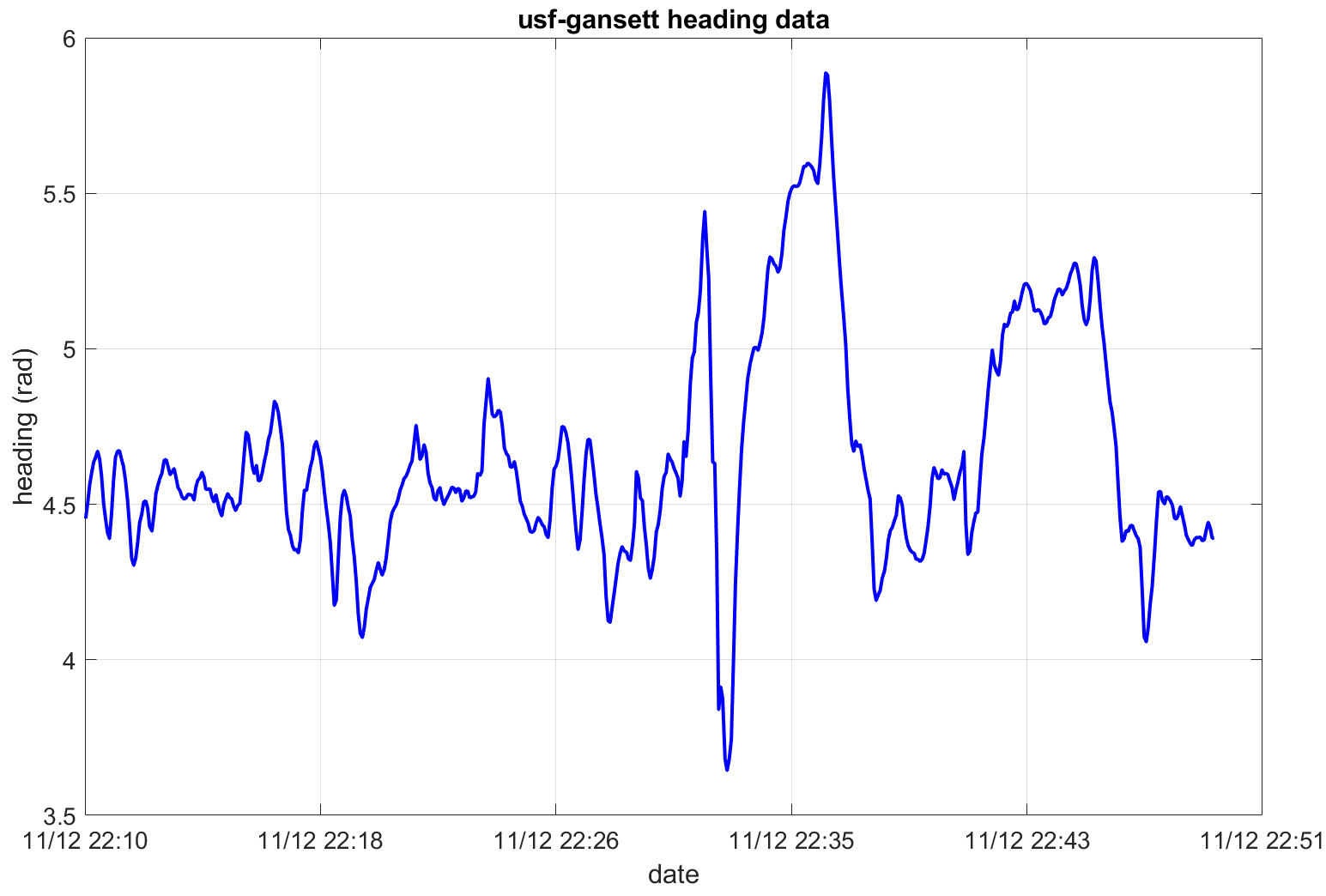}}
         \caption{Glider-measured heading (m) from post-recovery DBD data.}
         \label{usf-gansett heading}
     \end{subfigure}

        \caption{Ground truth for the 2021 USF-Gansett deployment.}
        \label{usf-gansett ground truth data}
    \end{figure}

At November 12, 2021, 22:32 UTC, the glider USF-Gansett sharply rolled to starboard $47 \degree$ and pitches to $54 \degree$, settling back by 22:36 UTC to a roll of $11\degree - 15 \degree$ and normal pitches as shown in  Fig.~\ref{usf-gansett roll} and Fig.~\ref{usf-gansett pitch}. Heading changes during this time also varied by over $100 \degree$ as shown in Fig.~\ref{usf-gansett heading}. This abnormal roll persisted even though pitch and heading returns to normal afterwards.

\begin{figure}[ht]
     \centering

     \hfill
     \begin{subfigure}[b]{0.48\textwidth}
         \centerline{\includegraphics[width=\textwidth, height=5cm]{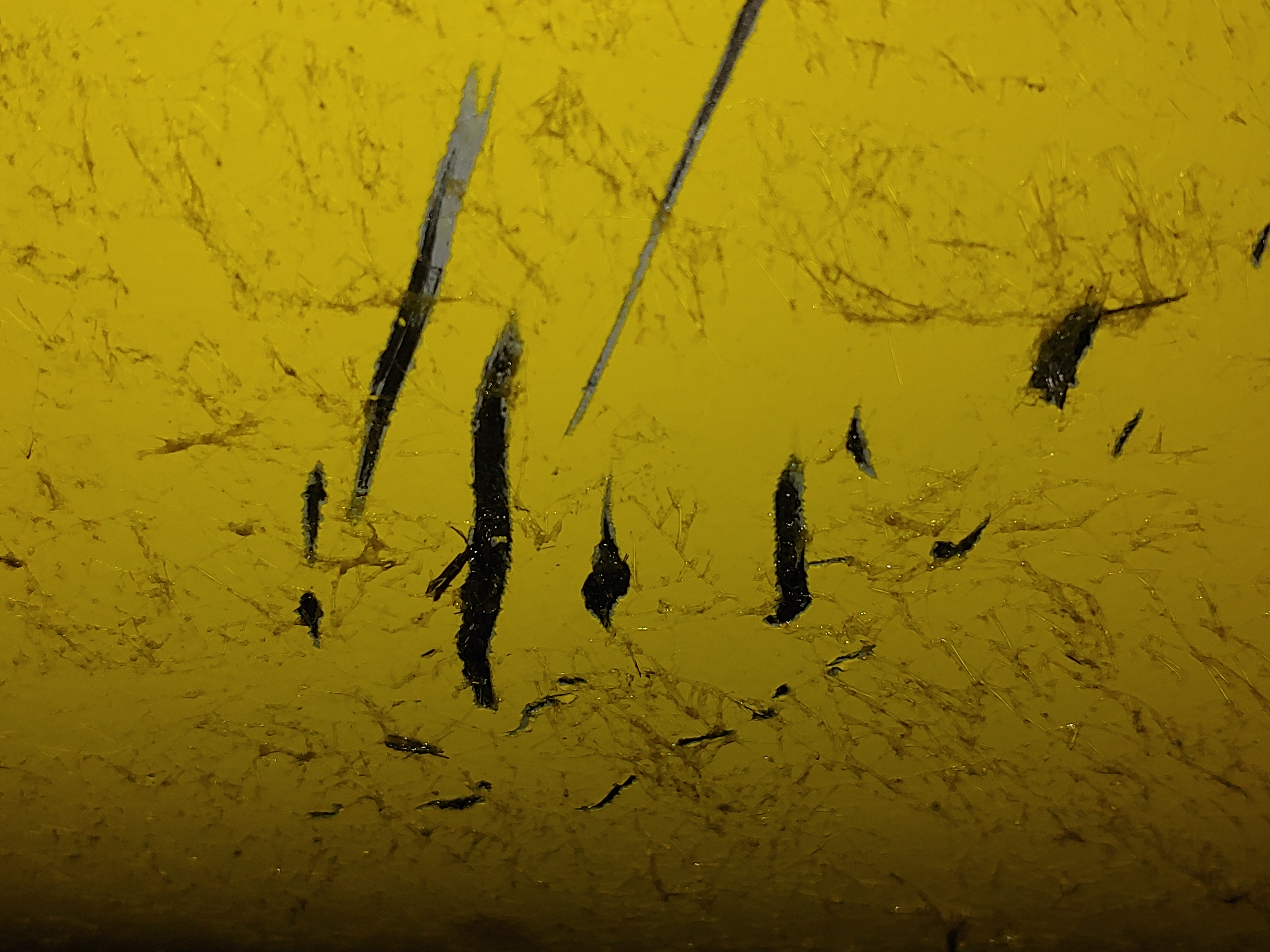}}
         \caption{Teeth marks on the aft hull of USF-Gansett}
         \label{usf-gansett teeth marks}
     \end{subfigure}
     
     \hfill
     \begin{subfigure}[b]{0.48\textwidth}
         \centerline{\includegraphics[width=\textwidth, height=5cm]{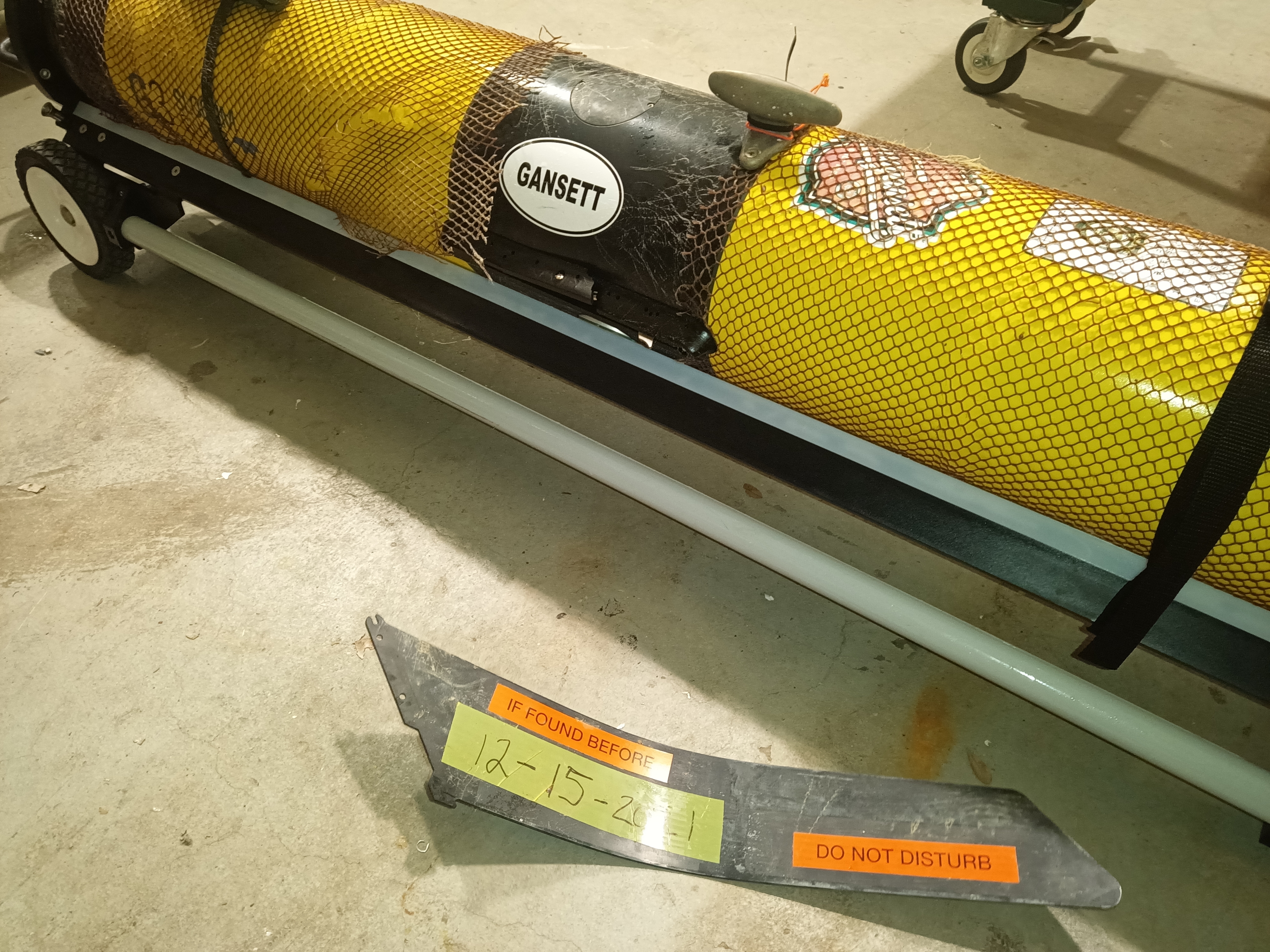}}
         \caption{Bent wing and cut netting upon recovery.}
         \label{usf-gansett bent wing}
     \end{subfigure}
     
        \caption{Post-recovery evidence of shark strike on USF-Gansett}
        \label{shark strike evidence}
    \end{figure}

Upon recovery, gouges resembling teeth marks are evident on the aft hull and science bay as shown in Fig.~\ref{usf-gansett teeth marks}. The arc of the marks span approximately 9 inches. The chord between the top and bottom ends of the aft hull markings is approximately 7.5 inches. The netting on the hull was cut in numerous areas, suggesting a serious shark strike. It is highly hypothesized that the bent starboard wing in Fig.~\ref{usf-gansett bent wing} is caused by the shark strike.

\begin{figure}[ht]
        \centerline{\includegraphics[width=0.48\textwidth]{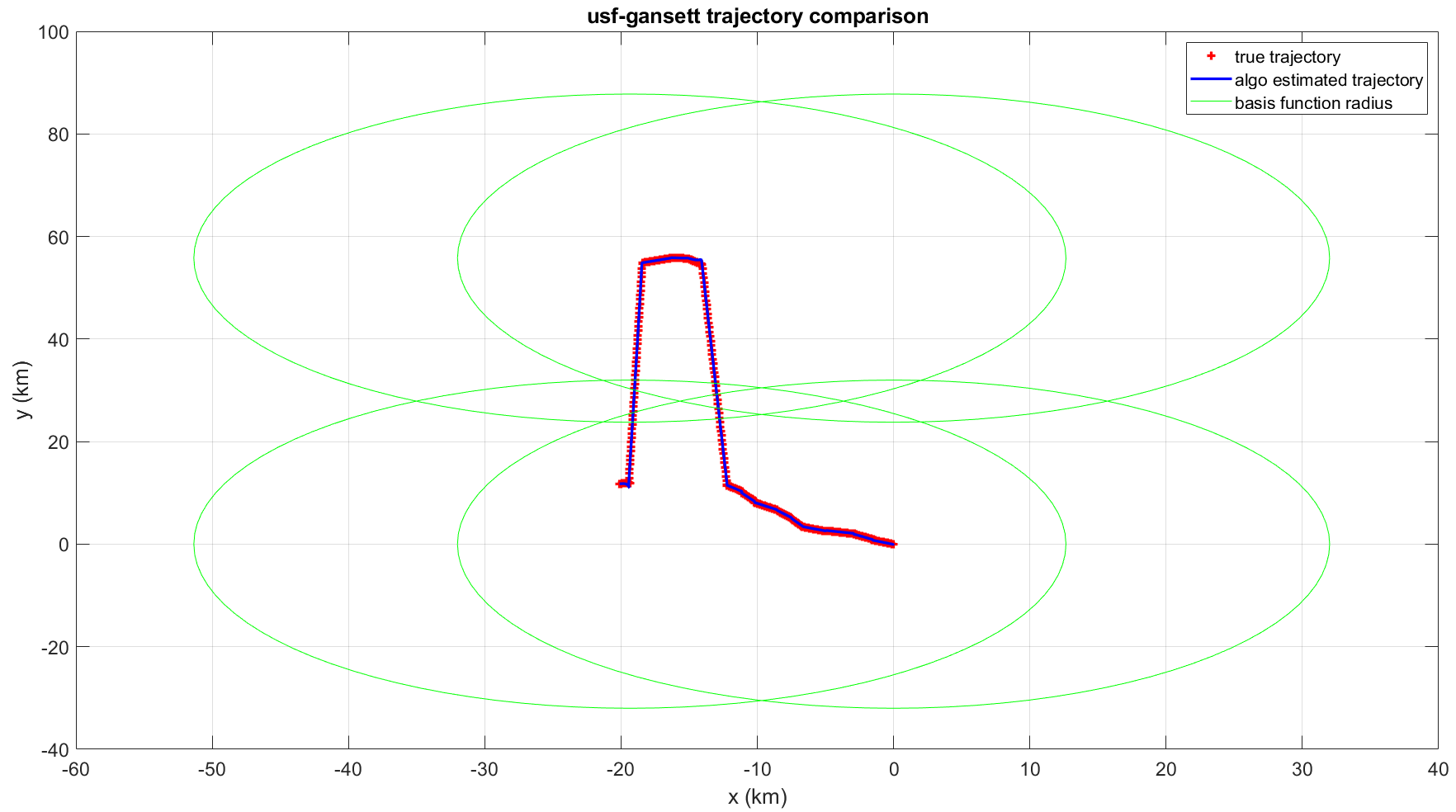}}
          \caption{Comparison of the estimated (blue) and true (red) trajectory  for the 2021 USF-Gansett deployment. The four green circles are the four basis functions covering the whole trajectory.}
        \label{usf-gansett trajectory comparison}
    \end{figure}

\begin{figure}[ht]
        \centerline{\includegraphics[width=0.48\textwidth]{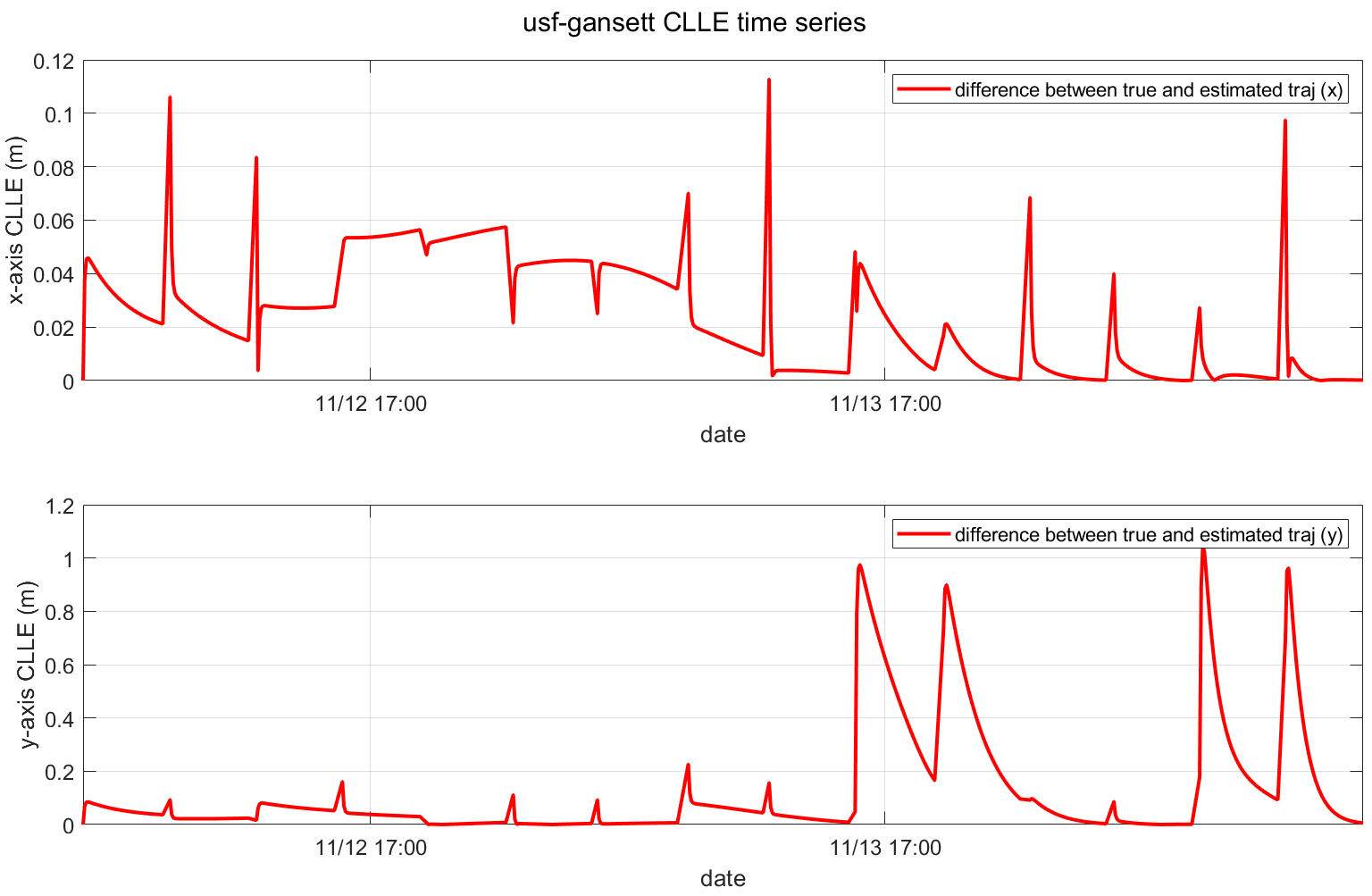}}
        \caption{CLLE (m) for the 2021 USF-Gansett deployment.}
        \label{usf-gansett CLLE}
    \end{figure}

\begin{figure}[ht]
     \centering

     \hfill
     \begin{subfigure}[b]{0.48\textwidth}
         \centerline{\includegraphics[width=\textwidth, height=3.5cm]{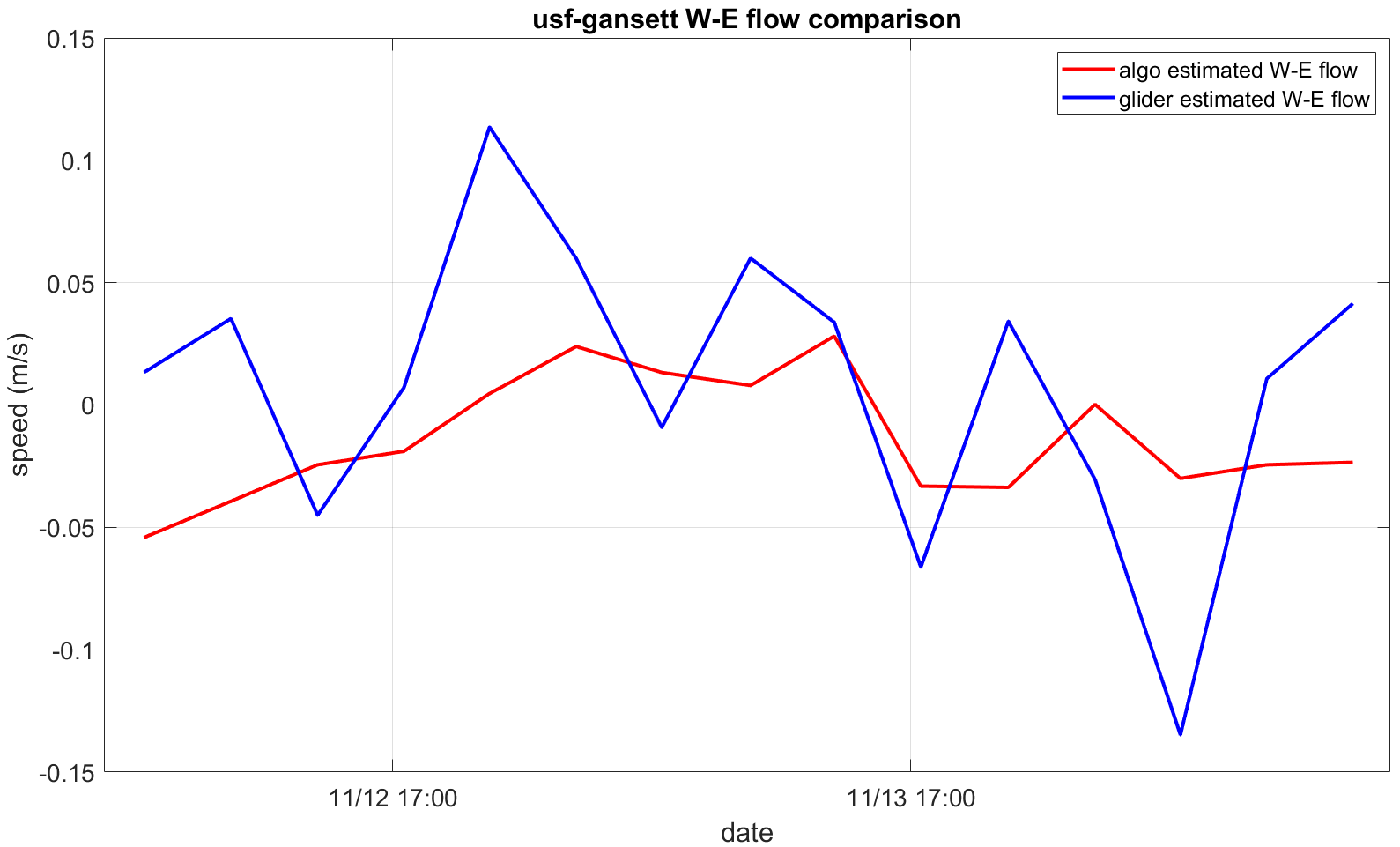}}
           \caption{W-E flow component.}
         \label{usf-gansett W-E flow}
     \end{subfigure}

     \hfill
     \begin{subfigure}[b]{0.48\textwidth}
         \centerline{\includegraphics[width=\textwidth, height=3.5cm]{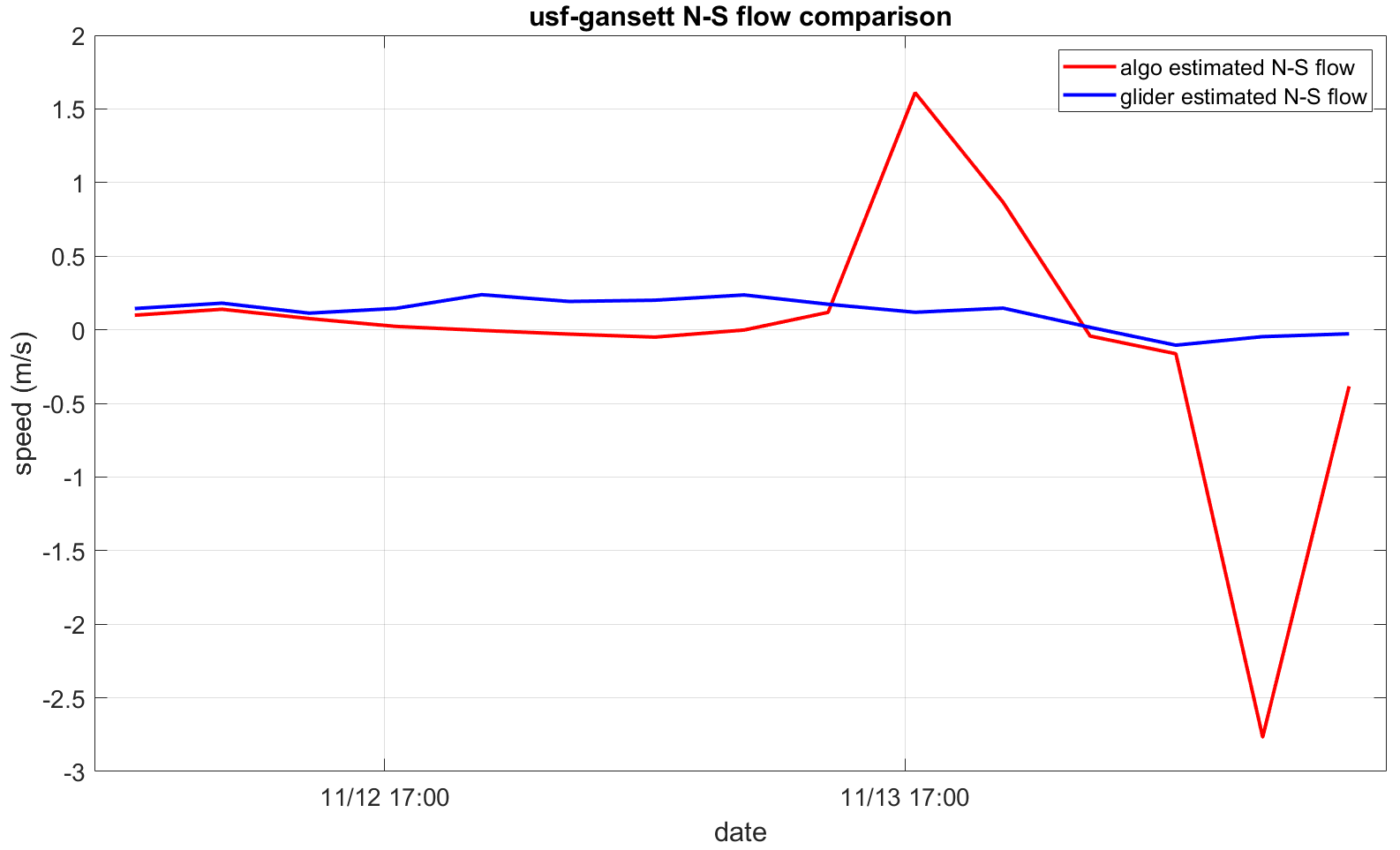}}  
           \caption{N-S flow component.}
         \label{usf-gansett N-S flow}
     \end{subfigure}
     
      \caption{Comparison of glider-estimated and algorithm-estimated W-E ($u$, upper) and N-S ($v$, lower) flow velocities for the 2021 USF-Gansett deployment.}
        \label{usf-gansett flow comparison}
    \end{figure}

\begin{figure}[ht]
        \centerline{\includegraphics[width=0.48\textwidth]{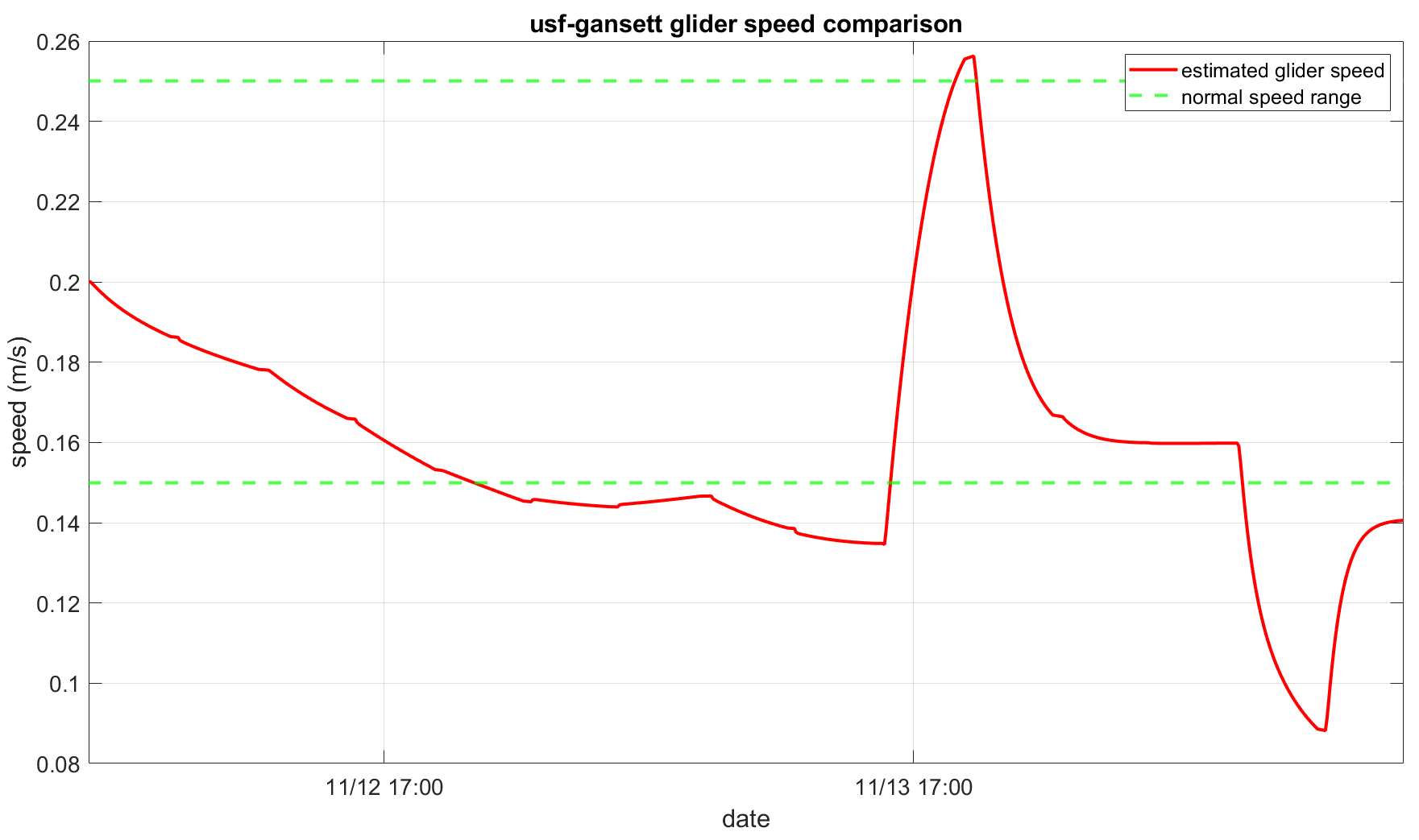}}
        \caption{Comparison of estimated glider speed (red) and normal speed range (green) for the 2021 USF-Gansett deployment.}
        \label{usf-gansett speed comparison}
    \end{figure}

Based on the DBD data, the estimated trajectory is close to the true trajectory as shown in Fig.~\ref{usf-gansett trajectory comparison}. From quantitative analysis in Fig.~\ref{usf-gansett CLLE}, the maximum CLLE $1.1m$ is small enough to conclude the CLLE convergence. We follow the same process of evaluating flow estimation error in Section \ref{franklin}. As shown in Fig.~\ref{usf-gansett flow comparison}, the small flow estimation error suggests that the detection result can be trusted. As shown in Fig.~\ref{usf-gansett speed comparison}, the estimated glider speed dropped out of the normal speed range (green dot line) at around November 12, 2021, 22:00 UTC, followed by radical speed changes that match the persistent roll change in the glider team's report. The timestamp when the anomaly is detected by the algorithm corresponds to the timestamp hypothesised by the glider team.

\subsubsection{USF-Stella}
\begin{figure}[ht]
     \centering

     \hfill
     \begin{subfigure}[b]{0.48\textwidth}
         \centerline{\includegraphics[width=\textwidth, height=3.5cm]{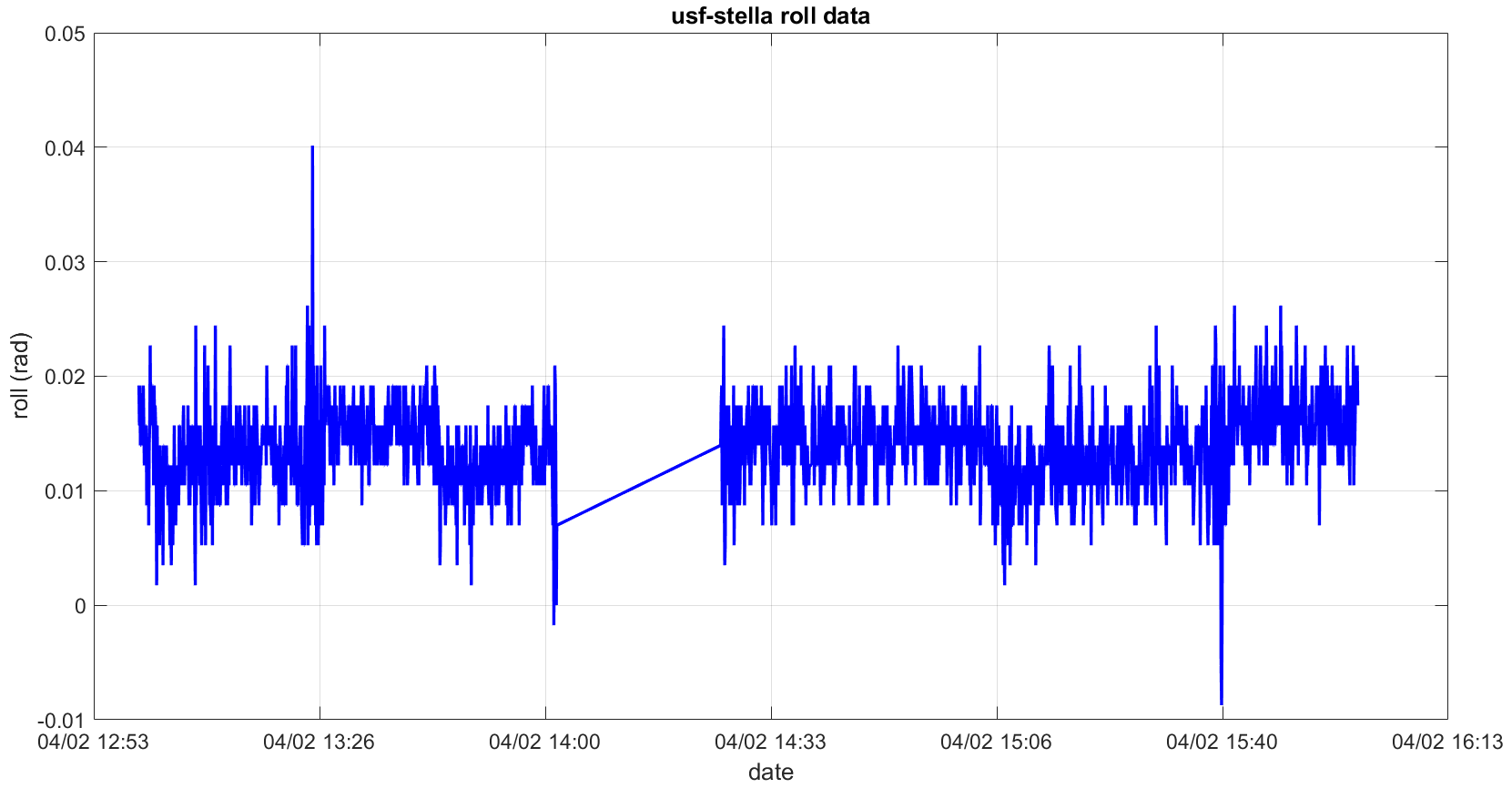}}
         \caption{Glider-measured roll (rad) from post-recovery DBD data.}
         \label{usf-stella roll}
     \end{subfigure}

    \hfill
     \begin{subfigure}[b]{0.48\textwidth}
         \centerline{\includegraphics[width=\textwidth, height=3.2cm]{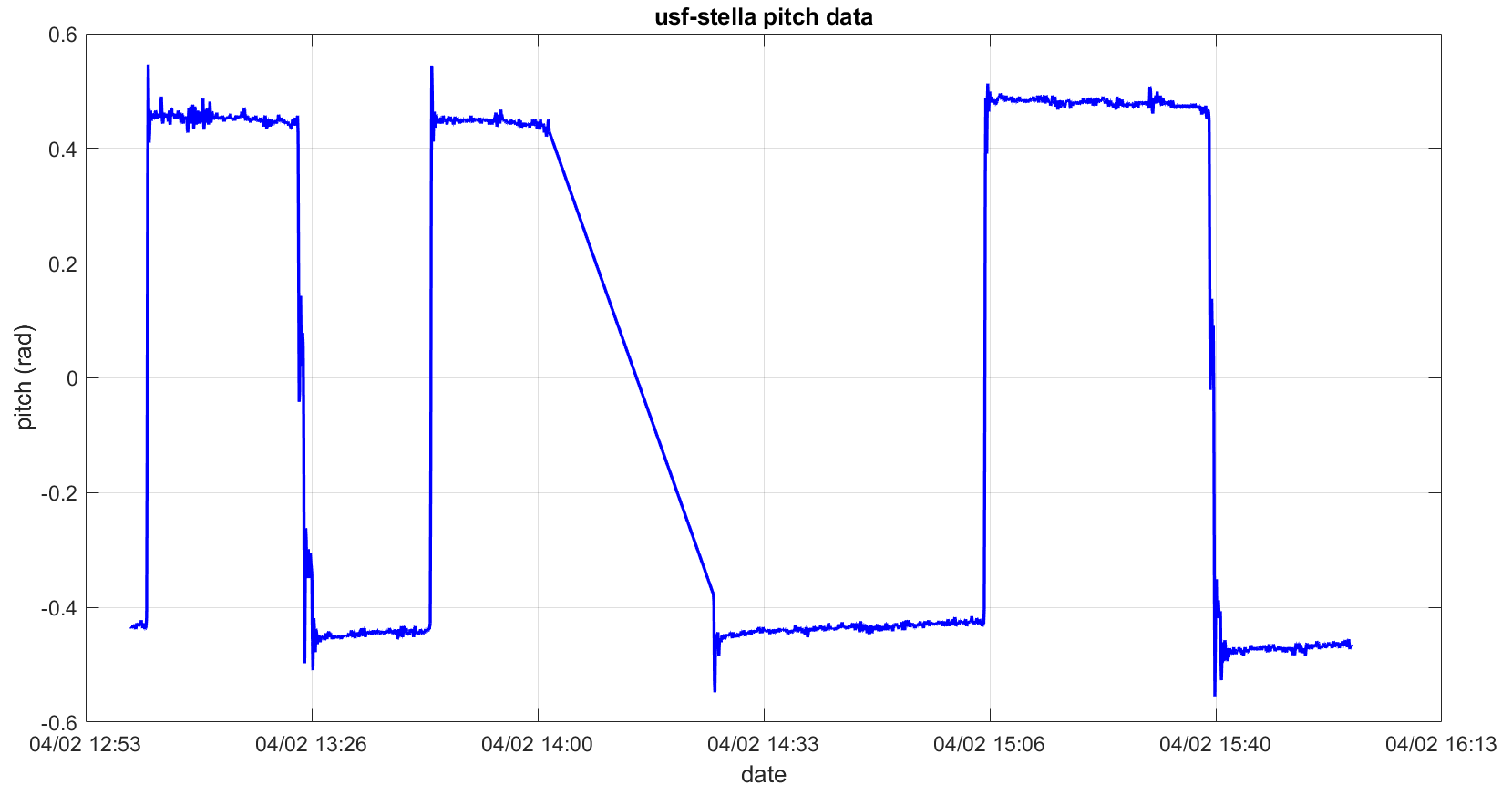}}
         \caption{Glider-measured pitch (rad) from post-recovery DBD data.}
         \label{usf-stella pitch}
     \end{subfigure}

    \caption{ground truth for the 2023 USF-Stella deployment.}
    \label{usf-stella ground truth data}
  \end{figure}

\begin{figure}[ht]
        \centerline{\includegraphics[width=0.4\textwidth, height=5cm, angle=270]{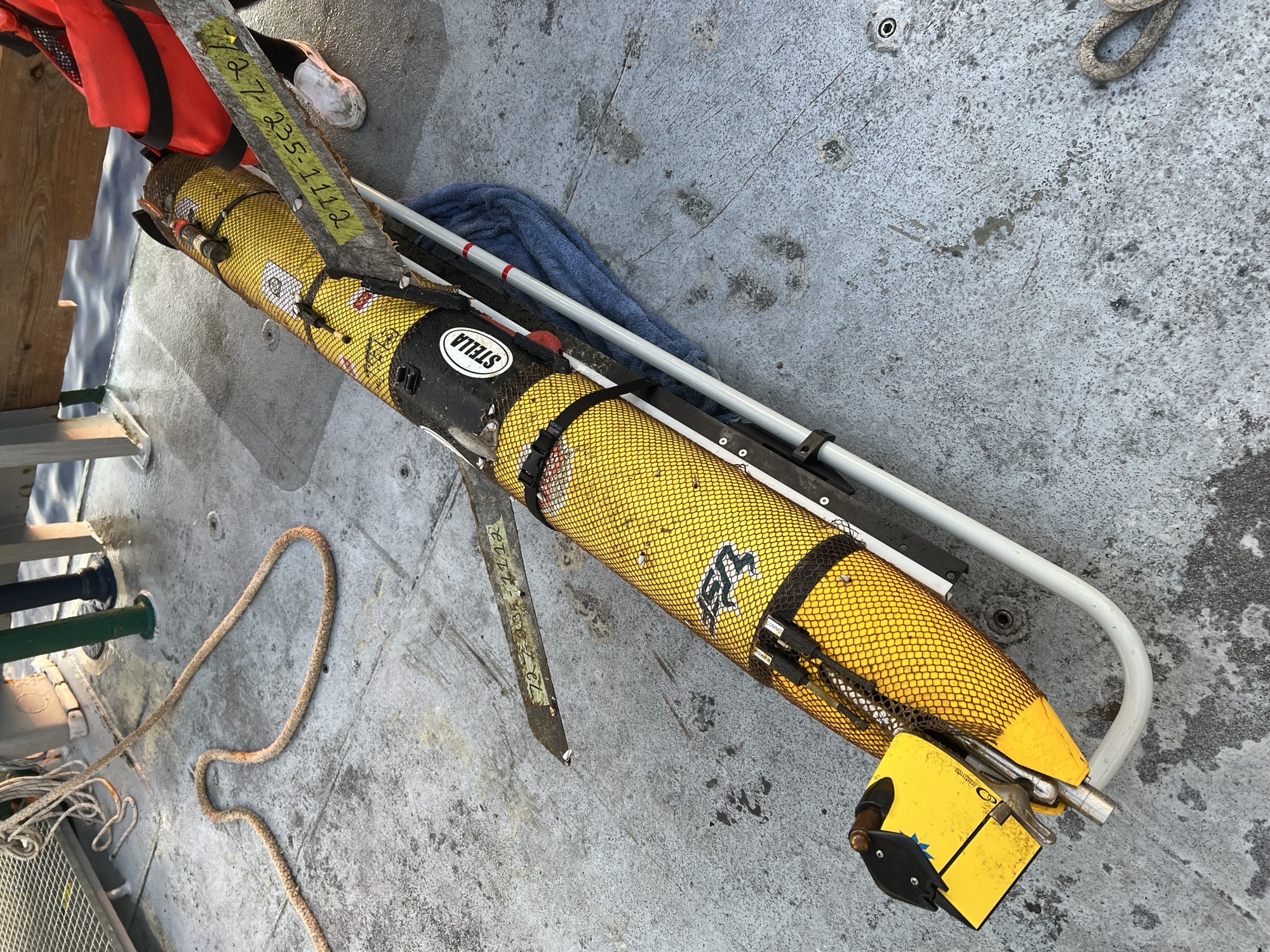}}
        \caption{Broken support rails of USF-Stella.}
        \label{usf-stella broken wing}
    \end{figure}

After performing hindcast analysis of the ground truth data as shown in Fig.~\ref{usf-stella ground truth data}, the glider team was certain that USF-Stella takes several hits during the deployment. At some point, the strike was serious enough that one of the wing support rails are broken, as shown in Fig.~\ref{usf-stella broken wing}.

\begin{figure}[ht]
        \centerline{\includegraphics[width=0.48\textwidth]{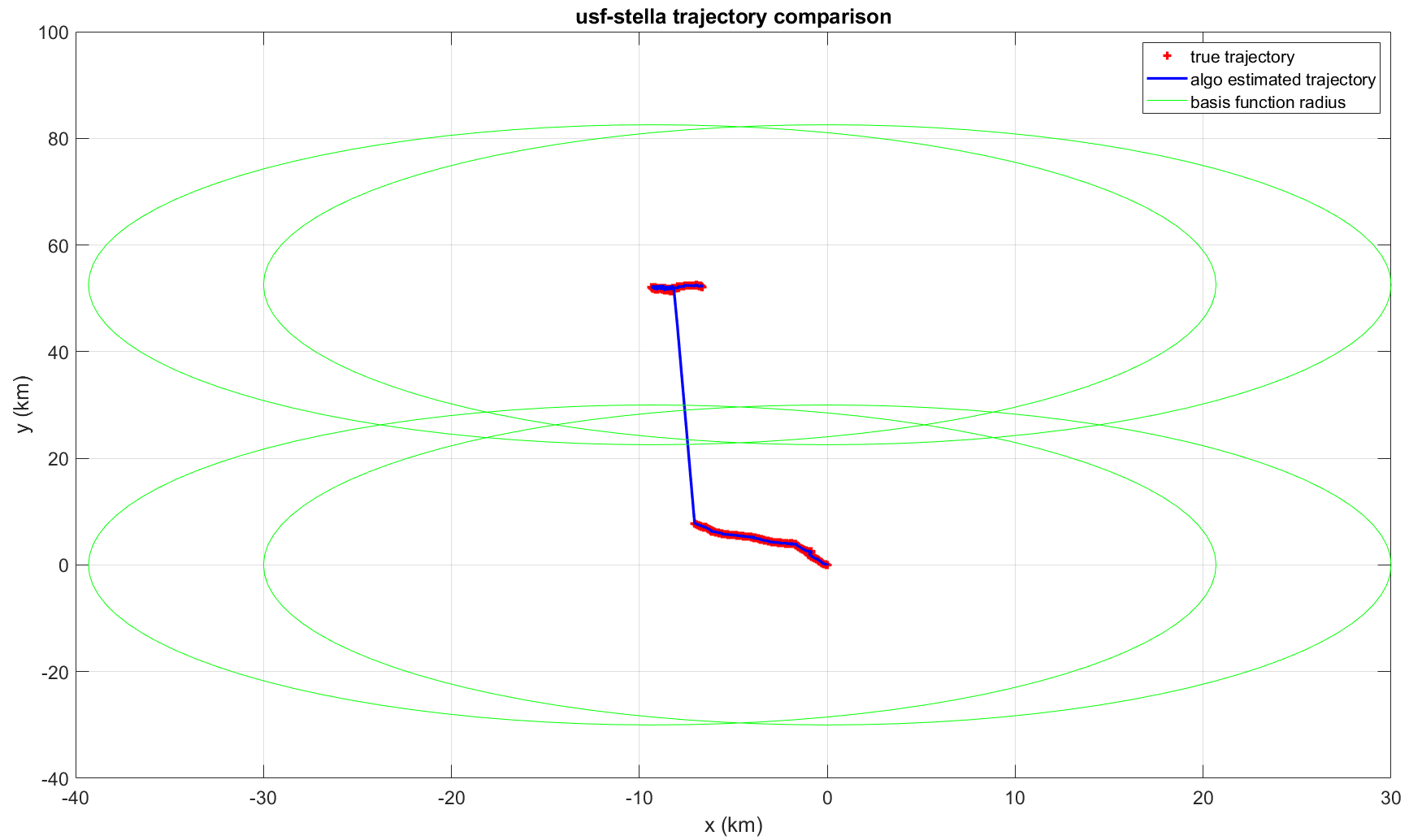}}
          \caption{Comparison of the estimated (blue) and true (red) trajectory  for the 2023 USF-Stella deployment. The four green circles are the four basis functions covering the whole trajectory.}
        \label{usf-stella trajectory comparison}
    \end{figure}

\begin{figure}[ht]
        \centerline{\includegraphics[width=0.48\textwidth]{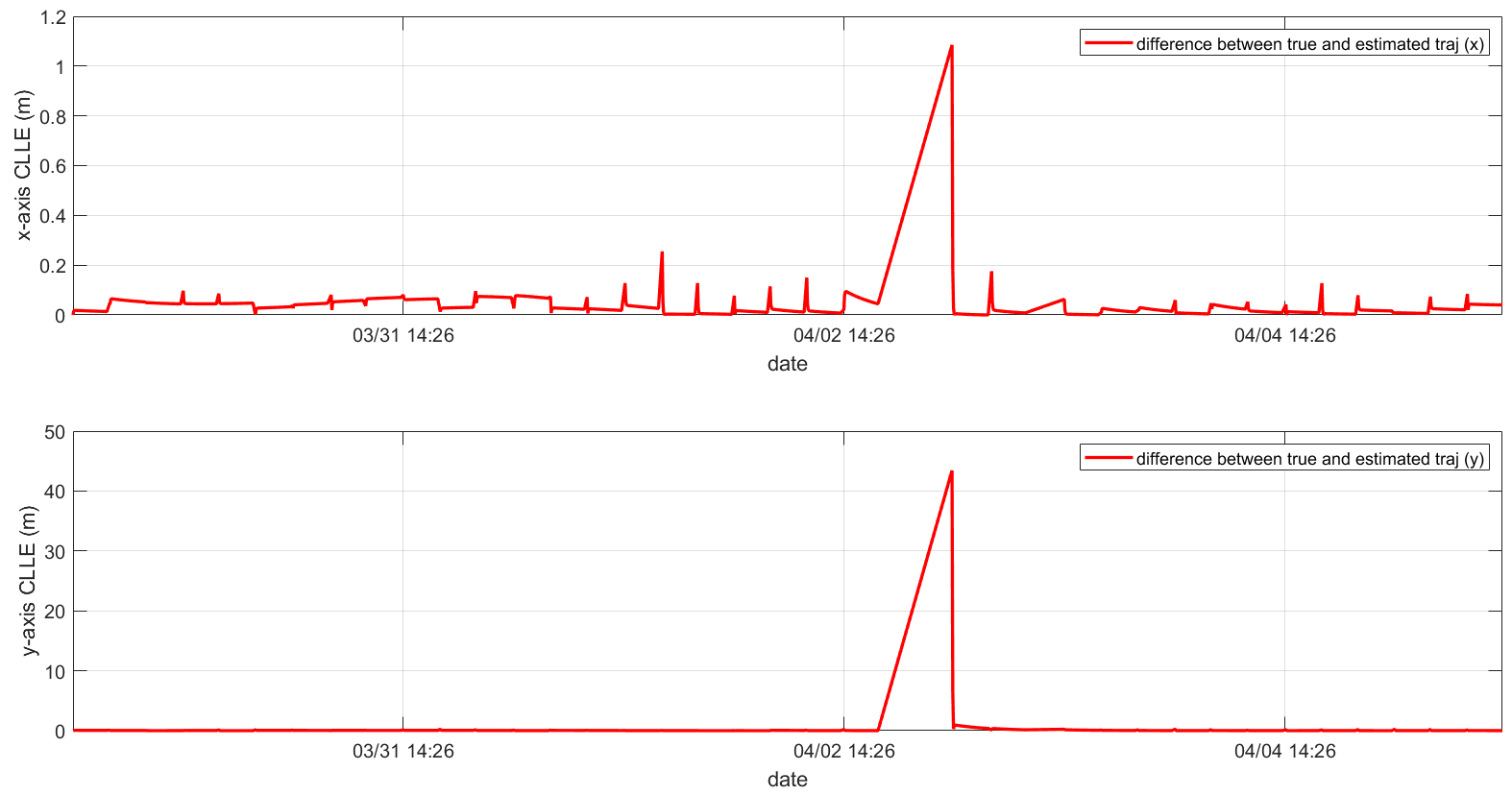}}
        \caption{CLLE (m) for the 2023 USF-Stella deployment.}
        \label{usf-stella CLLE}
    \end{figure}

\begin{figure}[ht]
     \centering

     \hfill
     \begin{subfigure}[b]{0.48\textwidth}
         \centerline{\includegraphics[width=\textwidth, height=3.5cm]{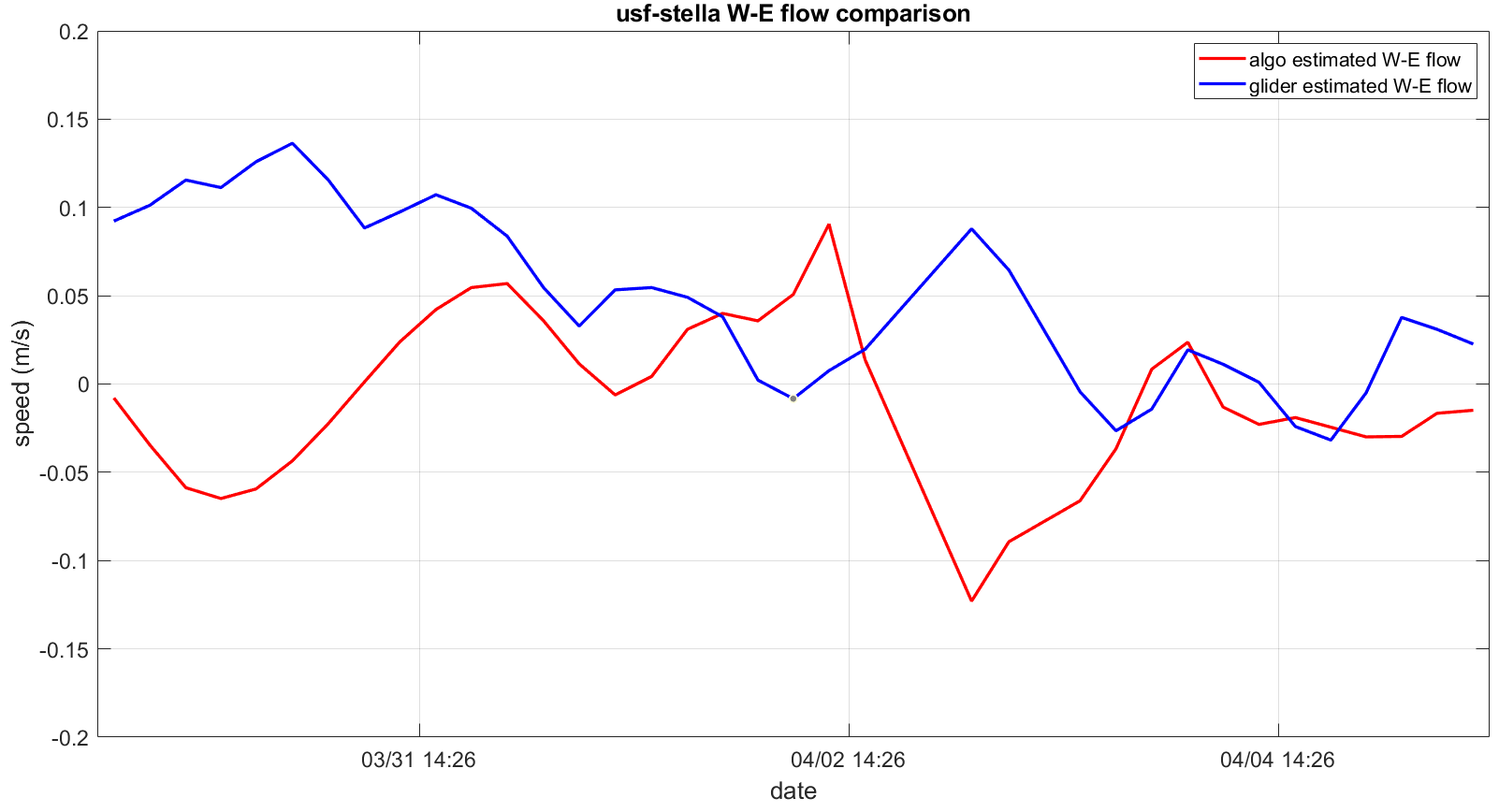}}
           \caption{W-E flow component.}
         \label{usf-stella W-E flow}
     \end{subfigure}

     \hfill
     \begin{subfigure}[b]{0.48\textwidth}
         \centerline{\includegraphics[width=\textwidth, height=3.5cm]{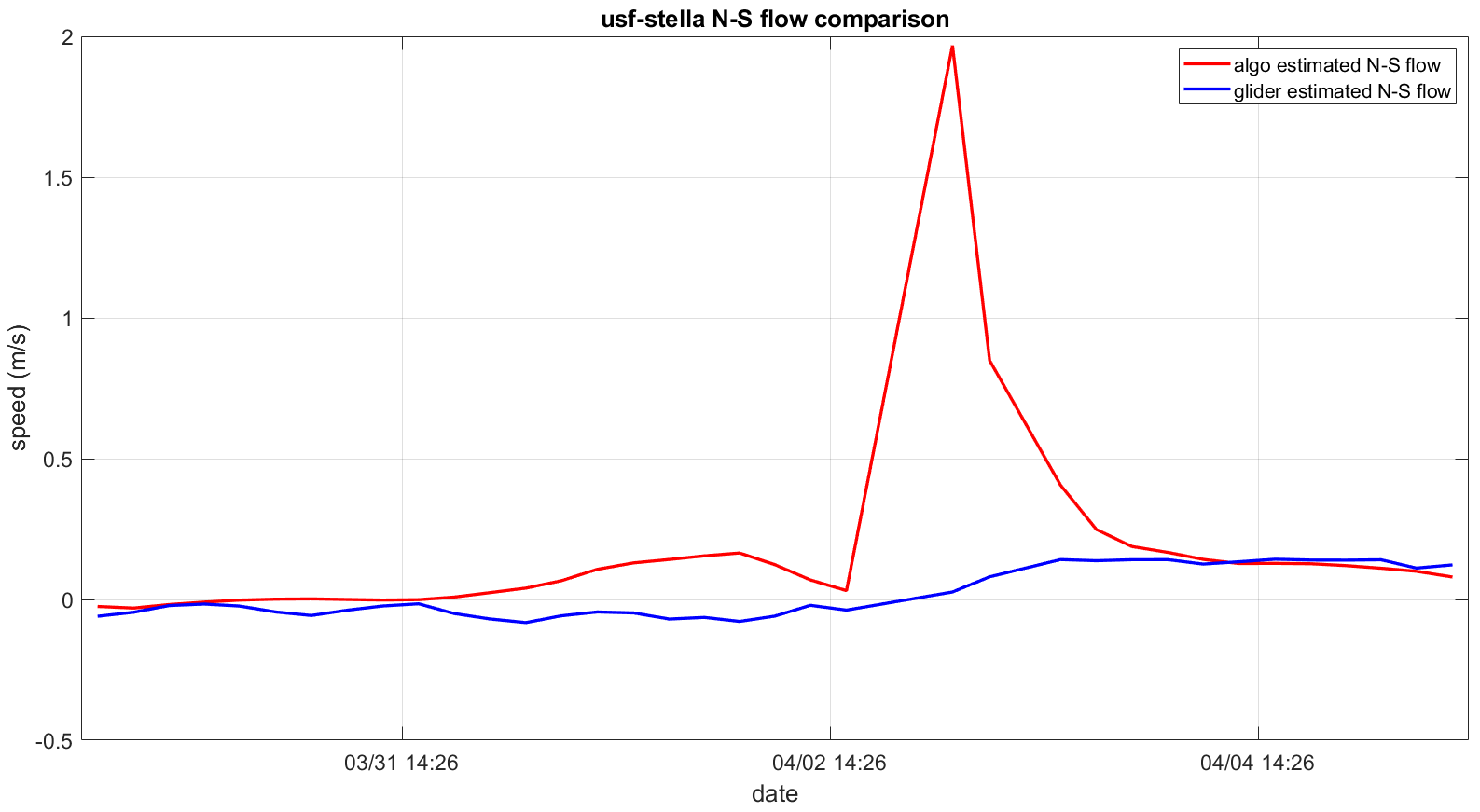}}  
           \caption{N-S flow component.}
         \label{usf-stella N-S flow}
     \end{subfigure}
     
      \caption{Comparison of glider-estimated and algorithm-estimated W-E ($u$, upper) and N-S ($v$, lower) flow velocities for the 2023 USF-Stella deployment.}
        \label{usf-stella flow comparison}
    \end{figure}

\begin{figure}[ht]
        \centerline{\includegraphics[width=0.48\textwidth]{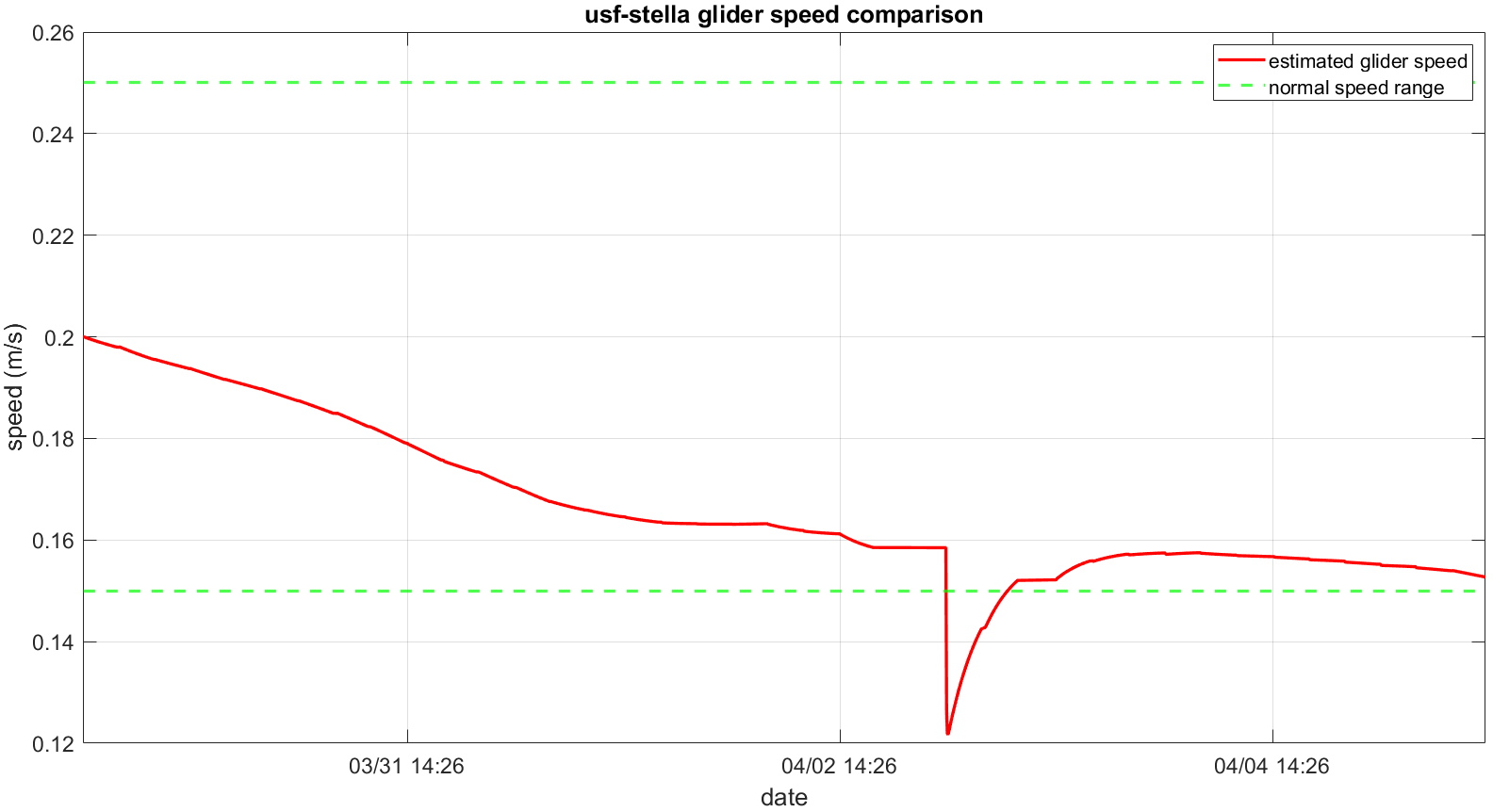}}
        \caption{Comparison of estimated glider speed (red) and normal speed range (green) for the 2023 USF-Stella deployment.}
        \label{usf-stella speed comparison}
    \end{figure}

Based on the DBD data, the algorithm-estimated trajectory is close to the true trajectory, as shown in Fig.~\ref{usf-stella trajectory comparison}. From quantitative analysis in Fig.~\ref{usf-stella CLLE}, the maximum CLLE $45m$ is small enough to conclude the CLLE convergence. We follow the same process of evaluating flow estimation error in Section \ref{franklin}. As shown in Fig.~\ref{usf-stella flow comparison}, the small flow estimation error suggests that the detection result can be trusted. As shown in Fig.~\ref{usf-stella speed comparison},  the estimated glider speed dropped out of the normal speed range (green dot line) at around April 02, 2023, 15:00 UTC. The anomaly timestamp detected by the algorithm corresponds to the timestamp in the glider team's report.

\section{Conclusion}
\label{conclusion}
In this paper, we apply an anomaly detection algorithm to four real glider missions supported by the Skidaway Institute of Oceanography in the University of Georgia and the University of South Florida. On one side of generality, the algorithm is capable of detecting anomalies like remora attachment and shark hit in diverse real-world deployments based on high-resolution DBD data. On the other side of real-time performance, we simulate the online detection on subsetted SBD data. It utilizes generic data of glider trajectory and heading angle to estimate glider speed and flow speed. Anomalies can be identified by comparing the estimated glider speed with the normal speed range. False alarms can be minimized by comparing the algorithm-estimated flow speed with the glider-estimated flow speed. The algorithm achieves real-time estimation through a model-based framework by continuously updating estimates based on ongoing deployment feedback. Future work will enhance estimation accuracy by incorporating large amount of  glider data into a data-driven framework. It is also worth taking into account the impact of the anomaly on the estimated flow speed, aiding in the process of determining false alarms.

\bibliography{IEEEabrv, reference.bib}
\bibliographystyle{IEEEtran}

\end{document}